\documentclass[10pt,twocolumn,letterpaper]{article}

\usepackage{wacv}              

\usepackage{graphicx}
\usepackage{amsmath}
\usepackage{amssymb}
\usepackage{booktabs}
\usepackage{multirow, multicol}
\usepackage[table]{xcolor}
\usepackage{graphicx}
\usepackage{adjustbox}
\usepackage{tikz}
\usepackage{tikz-3dplot}
\usepackage{pgfplots}
\usepgfplotslibrary{colormaps}
\usetikzlibrary{arrows.meta, bending}
\usetikzlibrary{decorations.markings}

\usepackage[pagebackref,breaklinks,colorlinks]{hyperref}

\usepackage[capitalize]{cleveref}
\crefname{section}{Sec.}{Secs.}
\Crefname{section}{Section}{Sections}
\Crefname{table}{Table}{Tables}
\crefname{table}{Tab.}{Tabs.}

\def\wacvPaperID{1598} 
\def\confName{WACV}
\def\confYear{2025}

\begin{document}

\title{Learning Action Hierarchies via Hybrid Geometric Diffusion}


\author{
Arjun Ramesh Kaushik \qquad Nalini K. Ratha \qquad Venu Govindaraju\\
University at Buffalo, SUNY\\
{\tt\small \{kaushik3, nratha, govind\}@buffalo.edu}
}
\maketitle

\begin{abstract}
Temporal action segmentation is a critical task in video understanding, where the goal is to assign action labels to each frame in a video. While recent advances leverage iterative refinement-based strategies, they fail to explicitly utilize the hierarchical nature of human actions. In this work, we propose HybridTAS - a novel framework that incorporates a hybrid of Euclidean and hyperbolic geometries into the denoising process of diffusion models to exploit the hierarchical structure of actions. Hyperbolic geometry naturally provides tree-like relationships between embeddings, enabling us to guide the action label denoising process in a coarse-to-fine manner: higher diffusion timesteps are influenced by abstract, high-level action categories (root nodes), while lower timesteps are refined using fine-grained action classes (leaf nodes). Extensive experiments on three benchmark datasets, GTEA, 50Salads, and Breakfast, demonstrate that our method achieves state-of-the-art performance, validating the effectiveness of hyperbolic-guided denoising for the temporal action segmentation task.
\end{abstract}

\section{Introduction}
\label{sec:intro}

\noindent Temporal Action Segmentation (TAS) aims to assign an action label to every frame in an untrimmed video, enabling fine-grained understanding of complex human activities. This task is crucial for applications such as human-computer interaction, video surveillance, and robotic perception. Despite recent progress, achieving accurate segmentation remains challenging due to variability in action durations, temporal dependencies, and ambiguous transitions between actions.

Most existing approaches follow a refinement-based strategy, where a sequence of models iteratively improves frame-level predictions by leveraging contextual and temporal cues  \cite{GTRM, ASRF, Global2Local, SSTDA, DTL, HASR, fifa}. Recently, diffusion models have emerged as a promising direction in this space, demonstrating strong performance due to their ability to model complex temporal dynamics in a generative manner \cite{DiffAct, ActFusion}. However, current diffusion-based methods treat action labels as flat categories, ignoring the rich hierarchical structure often present in human activities. For example, high-level actions like ``prepare meal" can be decomposed into sub-actions such as ``cut vegetables", ``boil water", and ``stir ingredients".
To address this limitation, we propose a novel approach that incorporates a hybrid of Euclidean and hyperbolic geometry into the denoising process of diffusion models to model hierarchical relationships between actions. 

Hyperbolic geometry is inherently suited for representing tree-like structures due to its exponential growth property, making it a natural choice for representing hierarchy in datasets. Additionally, hyperbolic distances are also a natural measure of uncertainty and class boundaries \cite{HybImgSeg}. Our core insight is to guide the denoising trajectory in a coarse-to-fine manner, aligning the generative process with the action hierarchy. At higher diffusion timesteps (early in the generation process), the model is guided by coarse, high-level action categories (root nodes), and as the noise is reduced (towards lower timesteps), the model is increasingly influenced by fine-grained, low-level actions (leaf nodes).

We instantiate this idea through Euclidean and hyperbolic losses in the DiffAct framework \cite{DiffAct}. These hybrid losses are applied in two phases: \textit{Stabilization Phase} and \textit{Guidance Phase}. In the \textit{Stabilization Phase}, the model learns global representational embeddings of actions (referred to as action prototypes) that are optimized in hyperbolic space. Next, the \textit{Guidance Phase} controls and enforces radial outward movement of denoised embeddings towards their action prototypes. This structured guidance enables the model to align its diffusion trajectory along the geodesic path connecting the action prototype and the origin. We evaluate our approach on three widely used benchmarks for TAS: GTEA \cite{gtea}, 50Salads \cite{50salads}, and Breakfast \cite{breakfast}. Our method consistently outperforms the SOTA baselines, demonstrating the benefit of our hybrid geometry optimization. 

To summarize, our contributions are as follows: (1) We propose a novel hierarchical diffusion model for temporal action segmentation, which integrates a hybrid of Euclidean and hyperbolic geometry to capture action hierarchies. (2) We introduce a two-step optimization strategy using hyperbolic loss functions. The first step refines the action labels from coarse to fine levels of abstraction. (3) The second step enforces the model to align its denoising trajectory with the semantic hierarchy of actions. Additionally, our model surpasses SOTA works with fewer inference steps.

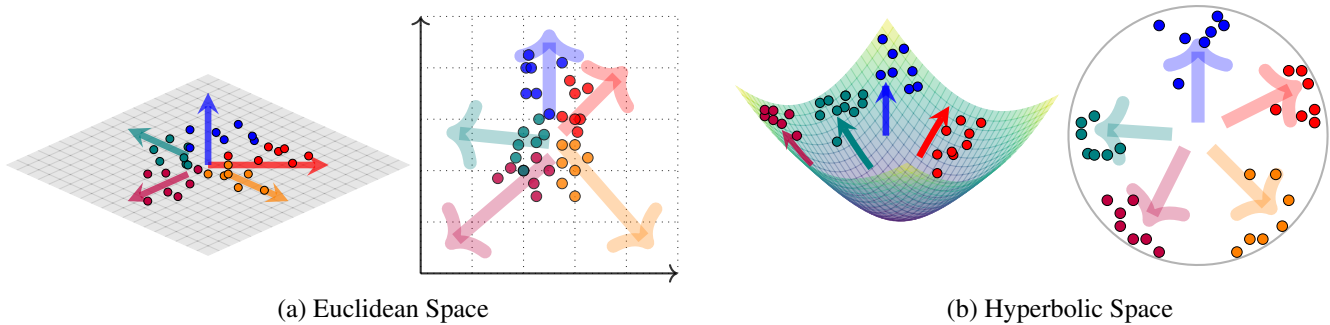
\begin{figure*}[!h]
    \centering
\begin{tikzpicture}
\node at (1,-0.5) {(a) Euclidean Space};
\node at (10,-0.5) {(b) Hyperbolic Space};
\begin{scope}[xshift=-4cm]
  \begin{axis}[
      view={45}{30},       
      axis lines=none,
      xmin=0, xmax=1.2,
      ymin=0, ymax=1.2,
      zmin=0, zmax=1.2,
      ticks=none,
      enlargelimits=false,
      clip=false,
      width=8cm, height=8cm,
      colormap/bone
    ]
    
    \addplot3[
      surf,
      opacity=0.1,
      domain=0:1,
      y domain=0:1,
      samples=20,
      samples y=20,
      shader=flat,
      colormap/bone
    ]
    ({x}, {y}, {0});

    \addplot3[thick, -stealth, red, opacity=0.7, line width = 2.5pt]
            coordinates {
        (0.5, 0.5, 0)
        (0.7, 0.7, 0)
        (0.8, 0.8, 0)};
    \addplot3[only marks, mark=*, mark size=1.5pt, draw = black, fill=red, mark options={line width=0.3pt}]
        coordinates {
            (0.5, 0.6, 0)
            (0.6, 0.65, 0)
            (0.7, 0.72, 0)
            (0.55, 0.75, 0)
            (0.7, 0.8, 0)
            (0.6, 0.75, 0)
            (0.6, 0.78, 0)
        };

    \addplot3[thick, -stealth, orange, opacity=0.7, line width = 2.5pt]
            coordinates {
        (0.6, 0.5, 0)
        (0.7, 0.5, 0)
        (0.9, 0.5, 0)};
    \addplot3[only marks, mark=*, mark size=1.5pt, draw = black, fill=orange, mark options={line width=0.3pt}]
        coordinates {
            (0.55, 0.55, 0)
            (0.55, 0.45, 0)
            (0.7, 0.45, 0)
            (0.6, 0.48, 0)
            (0.65, 0.55, 0)
            (0.6, 0.5, 0)
            (0.78, 0.5, 0)
        };

    \addplot3[thick, -stealth, purple, opacity=0.7, line width = 2.5pt]
            coordinates {
        (0.5, 0.4, 0)
        (0.5, 0.3, 0)
        (0.5, 0.1, 0)};
    \addplot3[only marks, mark=*, mark size=1.5pt, draw = black, fill=purple, mark options={line width=0.3pt}]
        coordinates {
            (0.55, 0.15, 0)
            (0.55, 0.25, 0)
            (0.44, 0.35, 0)
            (0.56, 0.35, 0)
            (0.5, 0.45, 0)
            (0.4, 0.3, 0)
            (0.6, 0.25, 0)
        };
    \addplot3[thick, stealth-, teal, opacity=0.7, line width = 2.5pt]
            coordinates {
        (0.1, 0.5, 0)
        (0.3, 0.5, 0)
        (0.4, 0.5, 0)};
    \addplot3[only marks, mark=*, mark size=1.5pt, draw = black, fill=teal, mark options={line width=0.3pt}]
        coordinates {
            (0.28, 0.45, 0)
            (0.35, 0.55, 0)
            (0.44, 0.45, 0)
            (0.36, 0.4, 0)
            (0.45, 0.45, 0)
            (0.4, 0.48, 0)
            (0.26, 0.55, 0)
        };
        
    \addplot3[thick, -stealth, blue, opacity=0.7, line width = 2.5pt]
            coordinates {
        (0.5, 0.5, 0)
        (0.3, 0.7, 0)
        (0.1, 0.9, 0)};
    \addplot3[only marks, mark=*, mark size=1.5pt, draw = black, fill=blue, mark options={line width=0.3pt}]
        coordinates {
            (0.48, 0.75, 0)
            (0.45, 0.77, 0)
            (0.34, 0.7, 0)
            (0.36, 0.55, 0)
            (0.35, 0.8, 0)
            (0.4, 0.68, 0)
            (0.26, 0.65, 0)
        };
    
  \end{axis}
\end{scope}

\begin{scope}[xshift=1.5cm]
    \begin{axis}[
      axis lines=left,
      xmin=0, xmax=1,
      ymin=0, ymax=1,
      ticks = none,
      enlargelimits=false,
      axis line style={->, thick},
      tick style={black},
      width=5cm, height=5cm,
      every axis x label/.style={at={(ticklabel* cs:1.05)}, anchor=west},
      every axis y label/.style={at={(ticklabel* cs:1.05)}, anchor=south},
      opacity = 0.8,
      grid=both,
      grid style={dotted, black},
    ]
    \addplot[only marks, mark=*, mark size=2pt, draw = black, fill=purple, mark options={line width=0.3pt}]
        coordinates {
            (0.4, 0.4)
            (0.45, 0.45)
            (0.5, 0.4)
            (0.45, 0.3)
            (0.35, 0.41)
            (0.44, 0.35)
            (0.3, 0.37)
        };

    \addplot[only marks, mark=*, mark size=2pt,draw = black, fill=orange, mark options={line width=0.3pt}]
        coordinates {
            (0.55, 0.43)
            (0.55, 0.35)
            (0.65, 0.45)
            (0.55, 0.5)
            (0.6, 0.5)
            (0.6, 0.4)
            (0.6, 0.3)
        };
    \addplot[only marks, mark=*, mark size=2pt, draw = black, fill=red, mark options={line width=0.3pt}]
        coordinates {
            (0.55, 0.61)
            (0.58, 0.75)
            (0.65, 0.72)
            (0.62, 0.6)
            (0.61, 0.55)
            (0.6, 0.6)
            (0.6, 0.7)
        };
    \addplot[only marks, mark=*, mark size=2pt, draw = black, fill=blue, mark options={line width=0.3pt}]
        coordinates {
            (0.41, 0.8)
            (0.42, 0.85)
            (0.55, 0.82)
            (0.45, 0.7)
            (0.41, 0.7)
            (0.43, 0.8)
            (0.5, 0.62)
        };

    \addplot[only marks, mark=*, mark size=2pt, draw = black, fill=teal, mark options={line width=0.3pt}]
        coordinates {
            (0.4, 0.4)
            (0.4, 0.55)
            (0.45, 0.51)
            (0.48, 0.54)
            (0.38, 0.5)
            (0.37, 0.45)
            (0.46, 0.6)
        };

    \addplot[->, thick, red, line width = 5pt, opacity=0.3] coordinates {(0.55,0.55) (0.8,0.8)};
    \addplot[->, thick, blue, line width = 5pt, opacity=0.3] coordinates {(0.5,0.6) (0.5,0.95)};
    \addplot[->, thick, teal, line width = 5pt, opacity=0.3] coordinates {(0.5,0.51) (0.1,0.55)};
    \addplot[->, thick, purple, line width = 5pt, opacity=0.3] coordinates {(0.51,0.45) (0.1,0.1)};
    \addplot[->, thick, orange, line width = 5pt, opacity=0.3] coordinates {(0.55,0.5) (0.9,0.1)};
    
  \end{axis}
\end{scope}
\begin{scope}[xshift=5cm, yshift=-0.75cm]
\begin{axis}[
    title={},
    view={45}{30},
    axis lines=center,
    colormap/viridis,
    enlargelimits=true,
    axis lines = none,
    ticks=none,
    domain=-2:2,
    samples=30,
    samples y=30,
    z buffer=sort,
    width=7cm,
    height=7cm,
]
    \addplot3[surf, opacity=0.35, shader=flat, colormap/viridis]
        ({x}, {y}, {sqrt(1 + x^2 + y^2)});


    \addplot3[thick, stealth-, red, opacity=1.0, line width = 2.5pt]
            coordinates {(-0.1, 0.7, 1.724)
        (-0.1, 0.9, 1.888)
        (-0.1, 1.1, 2.042)
        (-0.1, 1.3, 2.186)
        (-0.1, 1.5, 2.322)};
    \addplot3[only marks, mark=*, mark size=1.75pt, draw = black, fill=red, mark options={line width=0.3pt}]
        coordinates {
            (0.5, 0.5, 1.7)
            (0.5, 0.5, 2)
            (0.7, 0.7, 2)
            (0.7, 0.9, 2.5)
            (0.9, 0.9, 2.35)
            (0.8, 0.4, 2.5)
            (1.9, -0.3, 2.8)
            (1.9, -0.7, 2.8)
            (1.0, 1.1, 2.5)
        };

        \addplot3[only marks, mark=*, mark size=1.75pt, draw = black, fill=blue, mark options={line width=0.3pt}]
        coordinates {
            (0.8, -0.5, 3.7)
            (0.1, -0.5, 4)
            (0.9, -0.7, 3.5)
            (1.1, -0.7, 3.6)
            (0.6, -0.8, 3.7)
            (0.85, -1.4, 3.9)
            (0.6, -0.8, 4)
            (0.6, -0.5, 4)
        };
        \addplot3[thick, -stealth, blue, opacity=0.8, line width = 2.5pt]
            coordinates {
    (0, -0.3, 2.3664)
    (0, -0.3, 2.8328)
    (0, -0.3, 2.9992)
    (0, -0.3, 3.1656)
    (0, -0.3, 3.2320)};

    \addplot3[only marks, mark=*, mark size=1.75pt, draw = black, fill=teal, mark options={line width=0.3pt}]
        coordinates {
            (-1.5, 0.1, 1.7)
            (-1.5, 0.3, 2.2)
            (-1.5, 0.1, 2.2)
            (-1.1, 0.1, 2.3)
            (-1.3, 0.1, 2.5)
            (-1.3, 0.1, 2.5)
            (-0.9, 0.1, 2.5)
            (-0.8, 0.0, 2.7)
            (-1.9, 0.1, 2.0)
            (-1.9, 0.1, 2.2)
            (-1.7, 0.1, 2.3)
        };

     \addplot3[thick, stealth-, teal, opacity=1, line width = 2.5pt]
            coordinates {
    (-0.7, 0.1, 1.5207)
    (-0.9, 0.1, 1.6454)
    (-1.1, 0.1, 1.7870)
    (-1.3, 0.1, 1.9401)
    (-1.5, 0.1, 2.1028)
};

    \addplot3[only marks, mark=*, mark size=1.75pt, draw = black, fill=purple, mark options={line width=0.3pt}]
        coordinates {
            (-0.5, -2.1, 2.7)
            (-0.5, -2.3, 3.1)
            (-0.5, -2.5, 3)
            (-0.7, -2.5, 3)
            (-0.7, -2.5, 3.15)
            (-0.9, -2.5, 3)
            (-0.9, -2.5, 3.1)
        };
    \addplot3[thick, -stealth, purple, opacity=0.8, line width = 2pt]
            coordinates {
        (-0.5, -1.7, 2.124)
        (-0.5, -1.9, 2.3)
        (-0.5, -2.1, 2.5)
        (-0.5, -2.3, 2.7)
        (-0.5, -2.5, 2.9)};

\end{axis}
\end{scope}

\begin{scope}[xshift=9.25cm, yshift = -0.75cm]
\begin{axis}[
    title={},
    axis equal image,
    axis lines=none,
    enlargelimits=false,
    xmin=-1.5, xmax=1.5,
    ymin=-1.5, ymax=1.5,
    ticks=none,
    width=6.75cm,
    height=6.757cm
]

    \addplot[domain=0:360, samples=100, thick, black, opacity=0.3] ({cos(x)}, {sin(x)});

    \addplot[only marks, mark=*, mark size=2pt, draw = black, fill=purple, mark options={line width=0.3pt}]
        coordinates {
            (-0.6, -0.6)
            (-0.5, -0.5)
            (-0.5, -0.8)
            (-0.3, -0.9)
            (-0.4, -0.8)
            (-0.7, -0.5)
            (-0.6, -0.7)
        };

    \addplot[only marks, mark=*, mark size=2pt,draw = black, fill=orange, mark options={line width=0.3pt}]
        coordinates {
            (0.65, -0.7)
            (0.4, -0.3)
            (0.5, -0.8)
            (0.3, -0.9)
            (0.4, -0.8)
            (0.7, -0.5)
            (0.6, -0.3)
        };
    \addplot[only marks, mark=*, mark size=2pt, draw = black, fill=red, mark options={line width=0.3pt}]
        coordinates {
            (0.9, 0.1)
            (0.8, 0.15)
            (0.9, 0.2)
            (0.85, 0.4)
            (0.8, 0.5)
            (0.7, 0.5)
            (0.6, 0.2)
        };
    \addplot[only marks, mark=*, mark size=2pt, draw = black, fill=blue, mark options={line width=0.3pt}]
        coordinates {
            (0.1, 0.8)
            (0.2, 0.85)
            (0.15, 0.92)
            (-0.15, 0.4)
            (-0.1, 0.75)
            (-0.3, 0.85)
            (0.05, 0.72)
        };

    \addplot[only marks, mark=*, mark size=2pt, draw = black, fill=teal, mark options={line width=0.3pt}]
        coordinates {
            (-0.9, 0.1)
            (-0.8, 0.15)
            (-0.9, 0.01)
            (-0.85, -0.14)
            (-0.8, -0.15)
            (-0.7, -0.15)
            (-0.6, -0.1)
        };

    \addplot[->, thick, red, line width = 5pt, opacity=0.3] coordinates {(0.2,0.1) (0.8,0.4)};
    \addplot[->, thick, blue, line width = 5pt, opacity=0.3] coordinates {(0,0.1) (0,0.75)};
    \addplot[->, thick, teal, line width = 5pt, opacity=0.3] coordinates {(-0.2,0.01) (-0.8,0.05)};
    \addplot[->, thick, purple, line width = 5pt, opacity=0.3] coordinates {(-0.1,-0.1) (-0.4,-0.7)};
    \addplot[->, thick, orange, line width = 5pt, opacity=0.3] coordinates {(0.1,-0.1) (0.6,-0.6)};

\end{axis}
\end{scope}
\end{tikzpicture}
    \caption{\textbf{Embeddings in Euclidean and Hyperbolic Space.} (a) In the Euclidean space, embeddings tend to cluster towards the origin, making it difficult to distinguish classes. (b) On the other hand, the hyperbolic space allows embeddings to spread due to its exponential distance growth. Additionally, hyperbolic geometry naturally provides information on hierarchy in data, class boundaries, and uncertainty in predictions \cite{HybImgSeg}.}
    \label{fig:emb_spread}
\end{figure*}

\section{Related Work}

\paragraph{Temporal Action Segmentation.} TAS aims to assign frame-wise action labels to video sequences \cite{MSTCN++, ASFormer, DiffAct, DTL, SSTDA, Global2Local, LTContext, UVAST, ASRF, HASR}. Early methods addressed this task using temporal sliding windows \cite{6247801, karaman2014fast} or grammar-based approaches \cite{breakfast, kuehne2017weaklysupervisedlearningactions} to incorporate hierarchical structure in actions. With the rise of deep learning, temporal convolutional networks \cite{lea2016temporalconvolutionalnetworksaction, MSTCN++}  and transformer-based models \cite{ASFormer, UVAST} have been introduced to model temporal dependencies. However, capturing long-range temporal relations in videos remains challenging. To address this, several works \cite{GTRM, ASRF, Global2Local, SSTDA, DTL, HASR, fifa} have proposed iterative refinement strategies that operate on top of TAS predictions to improve performance. More recently, DiffAct \cite{DiffAct} and ActFusion \cite{ActFusion} leverage a diffusion-based framework to iteratively denoise action label predictions conditioned on video features. ActFusion \cite{ActFusion} adopts the diffusion process with a novel anticipative masking strategy, aiming to jointly unify TAS and Long-Term Action Anticipation.

\paragraph{Diffusion Models.} Diffusion-based generative models \cite{sohldickstein2015deepunsupervisedlearningusing, Croitoru_2023, song2022denoisingdiffusionimplicitmodels, ho2020denoisingdiffusionprobabilisticmodels}, which have been theoretically unified with score-based approaches \cite{song2020generativemodelingestimatinggradients, song2020improvedtechniquestrainingscorebased, song2021scorebasedgenerativemodelingstochastic}, are widely recognized for their stable training dynamics and the absence of adversarial mechanisms typically required in generative learning. These models have demonstrated remarkable success across various domains, including image generation \cite{dhariwal2021diffusionmodelsbeatgans, rombach2022highresolutionimagesynthesislatent, bhunia2023personimagesynthesisdenoising, 10.1145/3580305.3599412}, natural language generation \cite{yu2023latentdiffusionenergybasedmodel}, text-to-image synthesis \cite{gu2022vectorquantizeddiffusionmodel, kim2022diffusioncliptextguideddiffusionmodels}, and audio generation \cite{leng2022binauralgradtwostageconditionaldiffusion, lam2022bddmbilateraldenoisingdiffusion}. Recent advancements have proposed gradient-based guidance techniques to further improve sampling efficiency \cite{pmlr-v202-dinh23a}. While diffusion models have been repurposed for several image understanding tasks, such as object detection \cite{chen2023diffusiondetdiffusionmodelobject} and semantic segmentation \cite{baranchuk2022labelefficientsemanticsegmentationdiffusion, amit2022segdiffimagesegmentationdiffusion}, their application to video-related problems remains relatively limited. Notable exceptions include works on video forecasting and infilling \cite{yang2022diffusionprobabilisticmodelingvideo, höppe2022diffusionmodelsvideoprediction, voleti2022mcvdmaskedconditionalvideo}, as well as recent efforts in video memorability prediction \cite{sweeney2022diffusingsurrogatedreamsvideo} and frequency-aware video captioning \cite{10.1609/aaai.v37i3.25484}. 

\paragraph{Hyperbolic Geometry.} Hyperbolic geometry has become a powerful tool for embedding hierarchical and tree-like structures with minimal distortion \cite{nickel2017poincareembeddingslearninghierarchical, chami2020treescontinuousembeddingsback, desa2018representationtradeoffshyperbolicembeddings, 10.1007/978-3-642-25878-7_34}. Since the introduction of Hyperbolic Neural Networks (HNN) \cite{ganea2018hyperbolicneuralnetworks}, hyperbolic space has been successfully integrated into diverse neural architectures, including convolutional networks \cite{shimizu2021hyperbolicneuralnetworks}, attention-based models \cite{gulcehre2018hyperbolicattentionnetworks}, graph neural networks \cite{liu2019hyperbolicgraphneuralnetworks, chami2019hyperbolicgraphconvolutionalneural}, and more recently, vision transformers \cite{ermolov2022hyperbolicvisiontransformerscombining}. Recent studies have also used the hyperbolic radius to capture uncertainty \cite{10254452, ermolov2022hyperbolicvisiontransformerscombining, franco2023hyperbolicselfpacedlearningselfsupervised, flaborea2023certainitsanomalous} and to model hierarchical relationships such as parent-child structures \cite{nickel2017poincareembeddingslearninghierarchical, tifrea2018poincareglovehyperbolicword, surís2021learningpredictabilityfuture, ermolov2022hyperbolicvisiontransformerscombining, HybImgSeg}. In image segmentation, hyperbolic geometry has gained traction due to its strengths in uncertainty modeling and hierarchical representations \cite{HybImgSeg, franco2024hyperbolicactivelearningsemantic, 10254452}.

Despite significant progress in hyperbolic deep learning, its application to Temporal Action Segmentation (TAS) remains unexplored. Notably, existing approaches overlook the hierarchical structure of actions in the latent space. To the best of our knowledge, this is the first work to leverage hyperbolic geometry to guide the denoising trajectory of diffusion models for addressing the TAS task.

\section{Background}
\label{sec:background}
\noindent In this section, we provide a summary of DiffAct \cite{DiffAct} to contextualize our contributions and introduce Hyperbolic Geometry ~\cite{nickel2017poincareembeddingslearninghierarchical, chami2020treescontinuousembeddingsback}. For a better understanding of Diffusion Models \cite{sohldickstein2015deepunsupervisedlearningusing, Croitoru_2023, song2022denoisingdiffusionimplicitmodels, ho2020denoisingdiffusionprobabilisticmodels}, we refer the reader to Sec. \ref{sec:appendix_bg} (Appendix).


\subsection{Diffusion Action Segmentation (DiffAct)}
\noindent Diffusion models approximate a data distribution by corrupting samples with Gaussian noise and learning to reverse this process through iterative denoising. At each timestep $t$, clean data $\mathbf{x}_0$ is transformed into a noisy version $\mathbf{x}_t$ via a variance schedule $\gamma(t)$, while a neural network $f(\mathbf{x}_t, t)$ is trained to predict the original signal using an L2 loss. During inference, the model begins from pure noise and progressively refines it into a coherent sample. In our setting, this corresponds to generating frame-wise action label sequences from Gaussian noise, conditioned on video features for temporal action segmentation.
\begin{equation}
    \mathbf{x}_t = \sqrt{\gamma(t)} \, \mathbf{x}_0 + \sqrt{1 - \gamma(t)} \, \boldsymbol{\epsilon}, \quad \boldsymbol{\epsilon} \sim \mathcal{N}(0, \mathbf{I})
\end{equation}

\noindent DiffAct \cite{DiffAct} introduces a generative formulation for temporal action segmentation by leveraging denoising diffusion probabilistic models (DDPMs). Instead of directly predicting labels, it treats segmentation as iterative denoising, where action sequences are generated from pure noise conditioned on video features. The encoder–decoder framework integrates masking strategies inspired by human priors (position, boundary, and relation), which guide the model to better localize and infer actions. During inference, the decoder refines noisy label sequences step-by-step, optionally skipping intermediate steps for efficiency, until producing the final action segmentation. For more details, refer to Sec. \ref{subsec:DiffAct}.

\subsection{Hyperbolic Geometry}
\paragraph{Hyperbolic Space.} Hyperbolic space is a Riemannian manifold characterized by constant negative curvature \cite{nickel2017poincareembeddingslearninghierarchical}. It admits several isometric models, among which the $n$-dimensional Hyperboloid model, defined on the hypersurface \( \mathbb{H}^n_c \), is the most fundamental. The Poincaré Ball model \( \mathbb{B}^n_c \) can be obtained by projecting this hyperboloid onto a space-like hyperplane.

\paragraph{Poincaré Ball Model.} An $n$-dimensional Poincaré Ball model with constant sectional curvature \( -c \) is defined as the Riemannian manifold \( (\mathbb{B}^n_c, g_c) \), where
\[
\mathbb{B}^n_c = \{ x \in \mathbb{R}^n \mid c \|x\|^2 < 1 \},
\]
and the Riemannian metric is given by
\[
g_c(x) = \lambda^2_c(x) \, I_n,
\]
where \( \lambda_c(x) = \frac{2}{1 - c\|x\|^2} \) is the conformal factor and \( I_n \) is the Euclidean metric tensor, where \( \|x\| \) denotes the Euclidean norm of \( x \).

\paragraph{Exponential Map.}
Let \( x \in \mathbb{B}^n_c \) and \( v \in T_x\mathbb{B}^n_c \), the exponential map \( \exp^c_x: T_x\mathbb{B}^n_c \to \mathbb{B}^n_c \) is given by, where $\oplus_c$ denotes the Möbius addition operator:
\begin{equation}
\label{eq:exp}
    \exp^c_x(v) = x \oplus_c  \frac{1}{\sqrt{c}} \tanh\left( \sqrt{c} \lambda_x^c \frac{\|v\|}{2} \right) [v]
\end{equation}

\paragraph{Distance Function.}
For \( x, y \in \mathbb{B}^n_c \), the hyperbolic distance is defined as:
\begin{equation}
\label{eq:dist}
d_\mathbb{B}(x, y) = \frac{2}{\sqrt{c}} \tanh^{-1} \left( \sqrt{c} \| -x \oplus_c y \| \right)
\end{equation}
The distance from any point to the origin in \( \mathbb{B}^n_c \) reflects its uncertainty \cite{HybImgSeg}. Specifically, the closer an embedding is to the center of the ball (origin), the higher the uncertainty.

\paragraph{Exterior Angle.}
We define the exterior angle between two points \( x \) and \( y \) in the Poincaré ball as the minimum angle between the axis of the tangent cone at \( x \) and the vector pointing toward \( y \) \cite{entailment_poincare}. Formally, the angle \( \theta(x, y) \) is as follows, where \( \langle x, y \rangle \) denotes the standard Euclidean inner product:
 {
\footnotesize
\begin{equation}
\label{eq:ext_angle}
\theta(x, y) = \cos^{-1} \left( 
\frac{
    \langle x, y \rangle (1 + \|x\|^2) - \|x\|^2(1 + \|y\|^2)
}{
    \|x\| \cdot \|x-y\| \sqrt{1 + \|x\|^2} \left( \|y\|^2 - 2 \langle x, y \rangle \right)
}
\right)
\end{equation}
}

\paragraph{Aperture.}
The aperture of a cone centered at point \( x \in \mathbb{B}^d \) is given by:
\begin{equation}
\label{eq:aperture}
    \alpha(x) = \arcsin\left( \frac{K(1 - \|x\|^2)}{\|x\|} \right)
\end{equation}
where \( K \) is a scalar hyperparameter (usually set to 0.1), and \( \|x\| \) is the Euclidean norm of \( x \). This aperture decreases as \( \|x\| \to 1 \), i.e., as \( x \) approaches the boundary of the Poincaré ball \cite{entailment_poincare}.

\section{Problem Formulation} Temporal Action Segmentation (TAS) involves assigning a sequence of action labels to each frame in a video, effectively classifying every input frame into one of several predefined action classes. Formally, let a video be represented as a sequence of frames \( \mathbf{F} = [F_1, F_2, \ldots, F_L] \) of length \( L \). TAS aims to predict a sequence of frame-wise action labels \( \mathbf{A} = [A_1, A_2, \ldots, A_L] \), where each \( A_i \) is a one-hot vector corresponding to one of the $C$ action classes. 

\section{Method}

\begin{figure}[!t]
    \centering
    \includegraphics[width=0.4\textwidth]{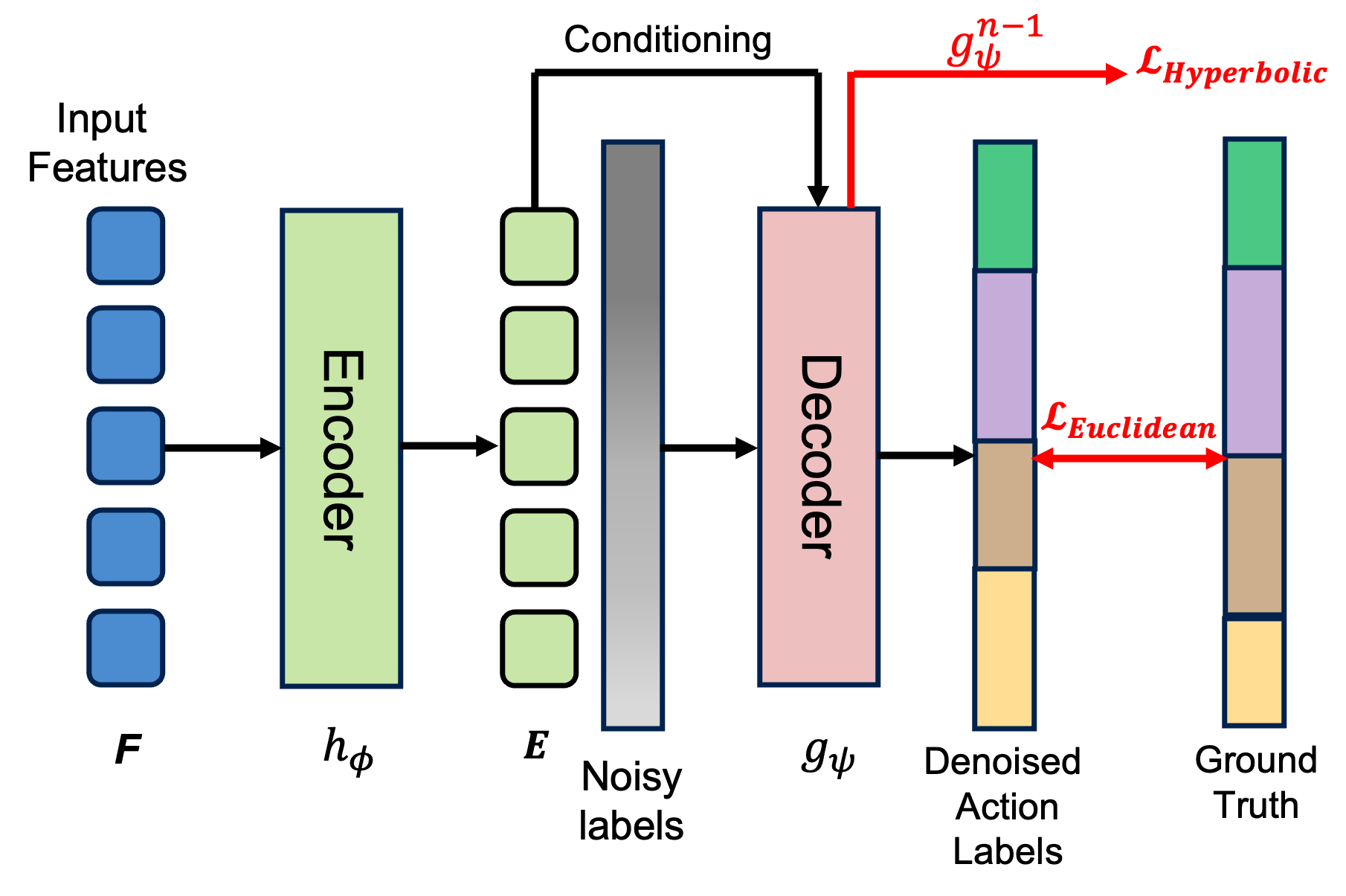}
    \caption{\textbf{Model Architecture}. Our architecture builds upon DiffAct~\cite{DiffAct}, but introduces a hybrid design that operates jointly in Euclidean and hyperbolic spaces to utilize hierarchical action relationships. The Euclidean losses are applied in the label space, acting on the predicted action probabilities, while the hyperbolic losses are computed in the embedding space, directly supervising the outputs from the decoder's final layer.}
    \label{fig:model}
\end{figure}

\noindent We present \textbf{HybridTAS}, a diffusion-based framework for temporal action segmentation that operates in both Euclidean and hyperbolic spaces. Our model builds upon DiffAct's \cite{DiffAct} diffusion architecture (see Sec.~\ref{subsec:DiffAct}), but with a critical shift: while DiffAct optimizes solely in label space, we optimize in both hyperbolic latent space and Euclidean label space (Fig. \ref{fig:model}). Our key innovation lies in structuring this trajectory to reflect the semantic hierarchy of action classes.

To this end, we define a unified training objective composed of four hyperbolic losses operating in the latent space:
(1) \textbf{Temporal Entailment Loss} to enforce temporal consistency across frames,
(2) \textbf{Prototype Margin Loss} for inter-class separation,  
(3) \textbf{Hyperbolic Push-Pull Loss} to reduce prediction uncertainty, and  
(4) \textbf{Geodesic Guidance Loss} to align denoising with hierarchical paths.  It is important to note that these losses operate on the decoder's final embeddings before outputting the action labels. On the other hand, we optimize in the label space using the standard \textbf{Cross-entropy Loss}.

Let $\mathbb{B}^d_c$ denote the $d$-dimensional Poincaré ball model with curvature parameter $c > 0$. For all hyperbolic losses, the embeddings are first projected into $\mathbb{B}^d_c$ using the manifold exponential map (Eq.~\ref{eq:exp}) prior to computing distances via $d_{\mathbb{B}}$ (Eq.~\ref{eq:dist}).

\subsection{Training Overview}
\noindent Training proceeds in two phases:  
(1) \textit{Stabilization Phase}, where action prototypes are learned dynamically, and  (2) \textit{Guidance Phase}, where prototypes are fixed and used to direct the denoising path. However, two losses optimize throughout both phases: \textbf{Cross-entropy Loss} and \textbf{Temporal Entailment Loss}. The losses, in this section, are defined as the optimization objective of the diffusion model at timestep $t$.

\paragraph{Cross-entropy Loss.} This is the standard cross-entropy for classification, minimizing the negative log-likelihood of the ground truth action labels for each frame. It operates directly in the Euclidean label space and is defined as:
\begin{equation}
\mathcal{L}_{\text{ce}} = \frac{1}{LC} \sum_{i=1}^L \sum_{c=1}^C -Y_{i,c} \log P_{i,c}
\end{equation}
where $i$ is the frame index and $c$ is the label index.

\paragraph{Temporal Entailment Loss.}
To ensure temporal consistency between adjacent denoised label embeddings, we impose a structural constraint based on angular entailment. Specifically, we interpret each frame embedding as semantically dependent on its predecessor. In hyperbolic space, this relation is captured by requiring the exterior angle (Eq. \ref{eq:ext_angle}) between consecutive embeddings to lie within the aperture (Eq. \ref{eq:aperture}) of the predecessor:
\begin{equation}
\mathcal{L}_{\text{entail}} = \frac{1}{L-1} \sum_{l=1}^{L-1} \max\left(0, \theta(x_l, x_{l+1}) - \alpha(x_l) \right)
\end{equation}
where $\theta(x_l, x_{l+1})$ is the exterior angle between $x_{l}$ and $x_{l+1}$. $\mathcal{L}_{\text{entail}}$ is not phase dependent and enforces temporal smoothness throughout training.

\begin{figure}[!b]
    \centering
    \includegraphics[width=0.4\textwidth]{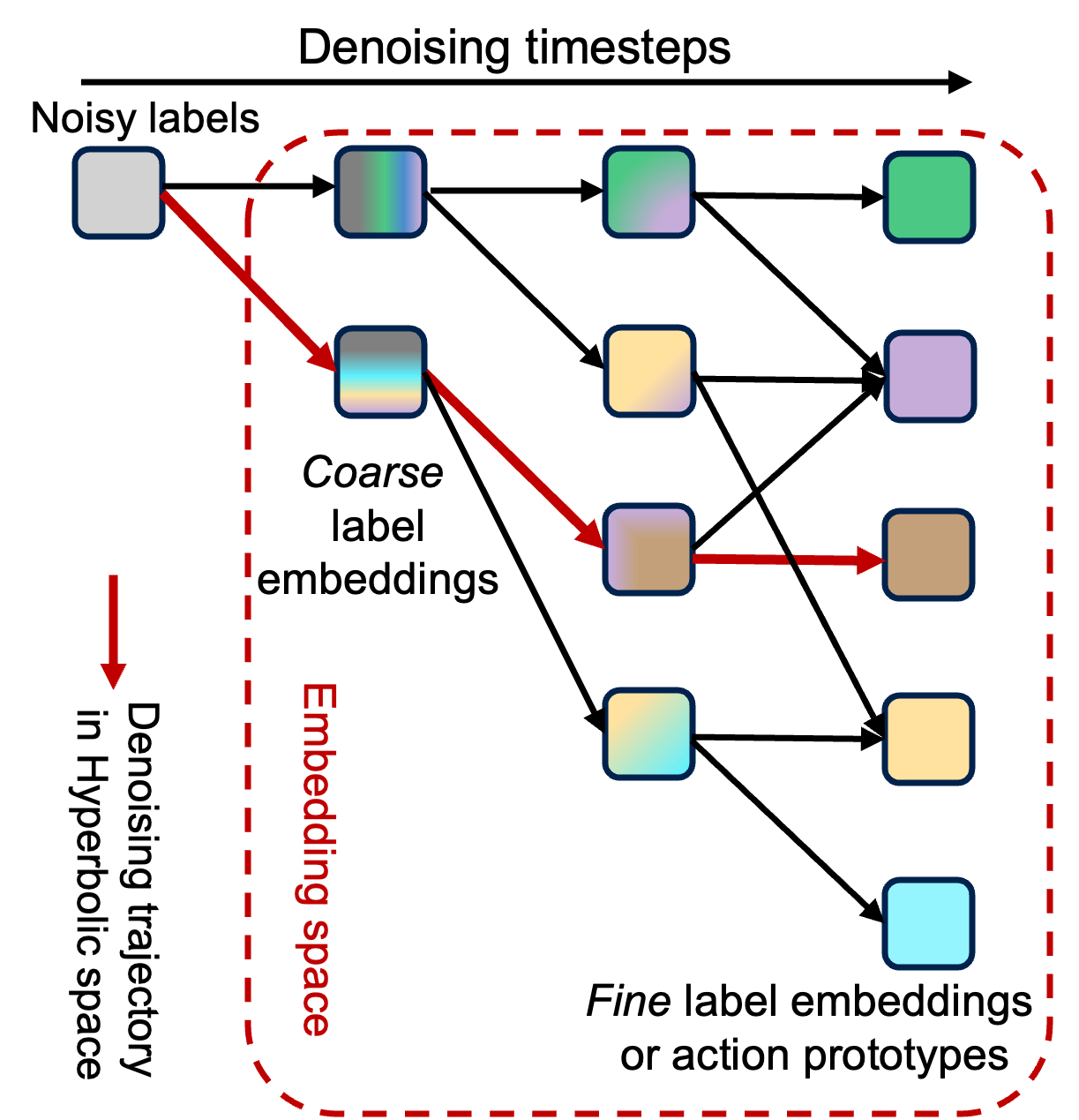}
    \caption{\textbf{Denoising trajectory in hyperbolic space.} Our hyperbolic loss functions guide the diffusion model to align its denoising trajectory along the geodesic between the origin and the target action prototype. This enforces the model to follow a hierarchical, coarse-to-fine progression in the label embedding space.}
    \label{fig:denoising}
\end{figure}

\begin{table*}[!t]
\centering
\renewcommand{\arraystretch}{1.3}
\resizebox{\textwidth}{!}{
\begin{tabular}{c|ccc|cc|c|ccc|cc|c|ccc|cc|c}
\toprule
\toprule
\multirow{2}{*}{\textbf{Method}} & \multicolumn{6}{c|}{\textbf{50 Salads \cite{50salads}}} & \multicolumn{6}{c|}{\textbf{Breakfast \cite{breakfast}}} & \multicolumn{6}{c}{\textbf{GTEA}\cite{gtea}} \\
\cline{2-19}
 & \textbf{F1@10} & \textbf{F1@25} & \textbf{F1@50} & \textbf{Edit} & \textbf{Acc} & \textbf{Avg} & \textbf{F1@10} & \textbf{F1@25} & \textbf{F1@50} & \textbf{Edit} & \textbf{Acc} & \textbf{Avg} & \textbf{F1@10} & \textbf{F1@25} & \textbf{F1@50} & \textbf{Edit} & \textbf{Acc} & \textbf{Avg} \\
\midrule
\textbf{MS-TCN++}\cite{MSTCN++} & 80.7 & 78.5 & 70.1 & 74.3 & 83.7 & 77.5 & 64.1 & 58.6 & 45.9 & 65.6 & 67.6 & 60.4 & 88.8 & 85.7 & 76.0 & 83.5 & 80.1 & 82.8 \\

\rowcolor{blue!10}
\textbf{SSTDA} \cite{SSTDA} & 83.0 & 81.5 & 73.8 & 75.8 & 83.2 & 79.5 & 75.0 & 69.1 & 55.2 & 73.7 & 70.2 & 68.6 & 90.0 & 89.1 & 78.0 & 86.2 & 79.8 & 84.6 \\

\textbf{GTRM} \cite{GTRM} & 75.4 & 72.8 & 63.9 & 67.5 & 82.6 & 72.4 & 57.5 & 54.0 & 43.3 & 58.7 & 65.0 & 55.7 & - & - & - & - & - & - \\

\rowcolor{blue!10}
\textbf{BCN} \cite{BCN} & 82.3 & 81.3 & 74.0 & 74.3 & 84.4 & 79.3 & 68.7 & 65.5 & 55.0 & 66.2 & 70.4 & 65.2 & 88.5 & 87.1 & 77.3 & 84.4 & 79.8 & 83.4 \\

\textbf{MTDA} \cite{MTDA} & 82.0 & 80.1 & 72.5 & 75.2 & 83.2 & 78.6 & 74.2 & 68.6 & 56.5 & 73.6 & 71.0 & 68.8 & 90.5 & 88.4 & 76.2 & 85.8 & 80.0 & 84.2 \\

\rowcolor{blue!10}
\textbf{Global2Local} \cite{Global2Local} & 80.3 & 78.0 & 69.8 & 73.4 & 82.2 & 76.7 & 74.9 & 69.0 & 55.2 & 73.3 & 70.7 & 68.6 & 89.9 & 87.3 & 75.8 & 84.6 & 78.5 & 83.2 \\

\textbf{HASR} \cite{HASR}& 86.6 & 85.7 & 78.5 & 81.0 & 83.9 & 83.1 & 74.7 & 69.5 & 57.0 & 71.9 & 69.4 & 68.5 & 90.9 & 88.6 & 76.4 & 87.5 & 78.7 & 84.4 \\

\rowcolor{blue!10}
\textbf{ASRF} \cite{ASRF} & 84.9 & 83.5 & 77.3 & 79.3 & 84.5 & 81.9 & 74.3 & 68.9 & 56.1 & 72.4 & 67.6 & 67.9 & 89.4 & 87.8 & 79.8 & 83.7 & 77.3 & 83.6 \\

\textbf{ASFormer} \cite{ASFormer} & 85.1 & 83.4 & 76.0 & 79.6 & 85.6 & 81.9 & 76.0 & 70.6 & 57.4 & 75.0 & 73.5 & 70.5 & 90.1 & 88.8 & 79.2 & 84.6 & 79.7 & 84.5 \\

\rowcolor{blue!10}
\textbf{UARL} \cite{UARL} & 85.3 & 83.5 & 77.8 & 78.2 & 84.1 & 81.8 & 65.2 & 59.4 & 47.4 & 66.2 & 67.8 & 61.2 & 92.7 & 91.5 & 82.8 & 88.1 & 79.6 & 86.9 \\

\textbf{DPRN} \cite{DPRN} & 87.8 & 86.3 & 79.4 & 82.0 & 87.2 & 84.5 & 75.6 & 70.5 & 57.6 & 75.1 & 71.7 & 70.1 & 92.9 & 92.0 & 82.9 & 90.9 & 82.0 & 88.1 \\

\rowcolor{blue!10}
\textbf{SEDT} \cite{SEDT} & 89.9 & 88.7 & 81.1 & 84.7 & 86.5 & 86.2 & - & - & - & - & - & - & 93.7 & 92.4 & 84.0 & 91.3 & 81.3 & 88.5 \\

\textbf{TCTr} \cite{TcTr} & 87.5 & 86.1 & 80.2 & 83.4 & 86.6 & 84.8 & 76.6 & 71.1 & 58.5 & 76.1 & 77.5 & 72.0 & 91.3 & 90.1 & 80.0 & 87.9 & 81.1 & 86.1 \\

\rowcolor{blue!10}
\textbf{FAMMSDTN} \cite{FAMMSDTN} & 86.2 & 84.4 & 77.9 & 79.9 & 86.4 & 83.0 & 78.5 & 72.9 & 60.2 & 77.5 & 74.8 & 72.8 & 91.6 & 90.9 & 80.9 & 88.3 & 80.7 & 86.5 \\

\textbf{DTL} \cite{DTL} & 87.1 & 85.7 & 78.5 & 80.5 & 86.9 & 83.7 & 78.8 & 74.5 & 62.9 & 77.7 & 75.8 & 73.9 & - & - & - & - & - & - \\

\rowcolor{blue!10}
\textbf{UVAST} \cite{UVAST} & 89.1 & 87.6 & 81.7 & 83.9 & 87.4 & 85.9 & 76.9 & 71.5 & 58.0 & 77.1 & 69.7 & 70.6 & 92.7 & 91.3 & 81.0 & 92.1 & 80.2 & 87.5 \\

\textbf{BrPrompt} \cite{BrPrompt} & 89.2 & 87.8 & 81.3 & 83.8 & 88.1 & 86.0 & - & - & - & - & - & - & 94.1 & 92.0 & 83.0 & 91.6 & 81.2 & 88.4 \\

\rowcolor{blue!10}
\textbf{MCFM} \cite{MCFM} & 90.6 & 89.5 & 84.2 & 84.6 & 90.3 & 87.8 & - & - & - & - & - & - & 91.8 & 91.2 & 80.8 & 88.0 & 80.5 & 86.5 \\

\textbf{LTContext} \cite{LTContext} & 89.4 & 87.7 & 82.0 & 83.2 & 87.7 & 86.0 & 77.6 & 72.6 & 60.1 & 77.0 & 74.2 & 72.3 & - & - & - & - & - & - \\

\rowcolor{blue!10}
\textbf{DiffAct }\cite{DiffAct} & 90.1 & 89.2 & 83.7 & 85.0 & 88.9 & 87.4 & 80.3 & 75.9 & 64.6 & 78.4 & 76.4 & 75.1 & 92.5 & 91.5 & 84.7 & 89.6 & 82.2 & 88.1 \\

\textbf{ActFusion }\cite{ActFusion} & 91.6 & 90.7 & 84.8 & 86.0 & 89.3 & 88.5 & 81.0 & 76.2 & 64.7 & 79.3 & 76.4 & 75.5 & 94.1 & 93.3 & 86.9 & 91.6 & 81.9 & 89.6 \\
\midrule
\rowcolor{gray!10}
\textbf{HybridTAS (Ours)}  & \textbf{92.8} & \textbf{91.8} & \textbf{88.4} & \textbf{89.4} & \textbf{90.6} & \textbf{90.6} & \textbf{82.8} & \textbf{77.9} & \textbf{68.1} & \textbf{81.1} & \textbf{80.2} & \textbf{78.0} & \textbf{97.0} & \textbf{97.0} & \textbf{90.8} & \textbf{95.2} & \textbf{83.5} & \textbf{92.7} \\

\bottomrule
\bottomrule
\end{tabular}}
\caption{\textbf{Quantitative Results.} Comparison of temporal action segmentation performance on GTEA \cite{gtea}, 50Salads\cite{50salads}, and Breakfast \cite{breakfast} datasets. Best results are denoted in bold.}
\label{tab:segmentation_results}
\end{table*}

Training consists of two phases: \textit{stabilization phase} and \textit{guidance phase}. The \textit{stabilization phase} is the early phase when the model learns the action prototype embeddings. Once the prototypes are stable, we begin the \textit{guidance phase} to refine predictions and provide the diffusion model with a hierarchical trajectory using the prototypes as targets. Recall that these losses operate on the final embeddings of the decoder.

\subsection{Step 1: Stabilization Phase}
\noindent We begin by initializing $C$ learnable embeddings for each action (action prototypes). These prototypes will serve as dynamic anchors in the \textit{stabilization phase} when the model will learn global action representations.

\paragraph{Prototype Margin Loss.}  
To avoid prototype collapse and promote discriminative representations, we introduce a margin-based repulsion loss:
\begin{equation}
    \mathcal{L}_{\text{margin}} = \frac{1}{C(C-1)} \sum_{i < j} \max \left(0, m - d_{\mathbb{B}}(z_i, z_j) \right)
\end{equation}
where \( z_i, z_j \in \mathbb{B}^d \) are action prototypes, \( m \) is a predefined margin, and \( d_{\mathbb{B}} \) is the Poincaré distance. This loss enforces a minimum separation between all prototype pairs, effectively distributing them across the manifold to preserve class-wise semantic boundaries. By maximizing inter-class distances in hyperbolic space, it enhances both alignment and separability, as illustrated in Fig.~\ref{fig:emb_spread}.

\paragraph{Hyperbolic Push-Pull Loss.}  
In hyperbolic space, an embedding’s distance from the origin naturally encodes prediction confidence, with points farther from the origin indicating lower uncertainty \cite{HybImgSeg}. We leverage this property by encouraging denoised embeddings to move outward (away from the origin) while simultaneously pulling them toward their corresponding prototypes. However, given that diffusion models denoise data in a coarse-to-fine manner \cite{coarse-to-fine, coarse-to-fine2}, applying a static distance-based loss is suboptimal. To address this, we introduce a timestep-aware objective that modulates the outward push using an exponential decay, allowing aggressive updates in early steps (lower values of $t$) and fine-grained alignment near convergence:
\begin{equation}
\small
    \mathcal{L}_{\text{pp}} = \frac{1}{N} \sum_{i=1}^{N} \left[\frac{d_{\mathbb{B}}(x_i, z_i)}{d_{\mathbb{B}}(O, x_i)} - d_{\mathbb{B}}(O, x_i) \cdot \exp\left(-\frac{t}{T}\right)\right]
\end{equation}
Here, \( x_i \) is the denoised embedding, \( z_i \) its corresponding prototype, \( O \) the origin of the Poincaré ball, and \( t/T \) represents the normalized timestep. The ratio term prevents prototypes from collapsing toward the origin, while the decay balances directional pull based on denoising progress.

The overall loss in the \textit{Stabilization Phase} can be summarized as:
{\small
\begin{equation}
    \mathcal{L}_{\text{stable}} = \lambda_{\text{ce}}\mathcal{L}_{\text{ce}} + \lambda_{\text{entail}}\mathcal{L}_{\text{entail}} + \lambda_{\text{margin}} \mathcal{L}_{\text{margin}} + \lambda_{\text{pp}} \mathcal{L}_{\text{pp}}
\end{equation}}

\subsection{Step 2: Guidance Phase}
\noindent Following the \textit{Stabilization Phase}, the action prototypes are fixed and serve as semantic anchors in the latent space. In the \textit{Guidance Phase}, the model learns to steer denoised embeddings toward their corresponding prototypes along geometrically meaningful paths (Fig. \ref{fig:denoising}).

\paragraph{Geodesic Guidance Loss.}
To align the diffusion trajectory with the shortest semantic path on the hyperbolic manifold, we propose a geodesic guidance loss:
\begin{equation}
\small
\mathcal{L}_{\text{gg}} = \frac{1}{N} \sum_{i=1}^{N} \left[ d_{\mathbb{B}}(O, z_i) - \left( d_{\mathbb{B}}(O, x_i) + d_{\mathbb{B}}(x_i, z_i) \right) \right]^2
\end{equation}
Here, $O$ denotes the origin, $z_i$ the target prototype, and $x_i$ the denoised embedding. The loss penalizes deviation from the geodesic triangle inequality, encouraging embeddings to evolve along the hyperbolic geodesic connecting the origin to the prototype via the intermediate point $x_i$. This enforces minimal deviation and nudges the trajectory toward semantically optimal paths.

The overall loss in the \textit{Guidance Phase} can be summarized as:
\begin{equation}
    \mathcal{L}_{\text{guidance}} = \lambda_{\text{ce}}\mathcal{L}_{\text{ce}} + \lambda_{\text{entail}}\mathcal{L}_{\text{entail}}  + \lambda_{\text{gg}} \mathcal{L}_{\text{gg}}
\end{equation}
\subsection{Total Loss}

\noindent The two-step training objective, where \textit{Step 1} utilizes $E_1$ epochs and $e$ be the current epoch, can be summarized as:
\begin{equation}
\mathcal{L}_{\text{total}} =  
\begin{cases}
\mathcal{L}_{\text{stable}}, & \text{if } e < E_1 \\
\mathcal{L}_{\text{guidance}}, & \text{if } e \geq E_1
\end{cases}
\end{equation}

The integration of Euclidean and hyperbolic losses is crucial: Euclidean loss ($\mathcal{L}_{\text{ce}}$) supervises the model's output predictions for local accuracy, while hyperbolic losses organize the embedding space for global hierarchy and separation. Euclidean losses alone cannot guarantee that the learned representations are hierarchically meaningful or robust to ambiguous cases, while hyperbolic losses alone cannot ensure frame-level accuracy. By jointly optimizing both sets of objectives, our model achieves accurate, temporally coherent, and boundary-aware action segmentation, underpinned by a geometry-aware, hierarchically structured representation space.

\section{Ablation Studies}
\noindent Extensive ablation studies are performed to validate the design choices in our method on the GTEA dataset \cite{gtea}.

\paragraph{Decaying function.} We study the impact of different decaying functions in $\mathcal{L}_{pp}$, which controls the radial outward movement of embeddings across timesteps in our model. As shown in Table~\ref{tab:decay}, the exponential decay function $e^{-x}$ yields the best performance across all metrics, achieving an average score of 92.7. This function sharply penalizes early errors while allowing more flexibility at later stages of the denoising process. In comparison, both the linear decay and cosine decay underperform slightly, with average scores of 91.2 and 92.0, respectively. These results demonstrate that sharper decay functions better align with the progressive nature of the diffusion trajectory and provide lower uncertainty in predictions.

\begin{table}[!h]

\renewcommand{\arraystretch}{1.3}
    \centering
    \resizebox{0.45\textwidth}{!}{
    \begin{tabular}{c|ccc|cc|c}
    \toprule
    \toprule
         \textbf{Decaying function} & \textbf{F1@10} & \textbf{F1@25} & \textbf{F1@50} & \textbf{Edit} & \textbf{Acc} & \textbf{Avg}  \\
    \midrule
\rowcolor{blue!10}
         $1-x$ & 95.2 & 94.4 & 90.6 & 92.8 & 82.9 & 91.2 \\
         $\frac{1}{2}(1 + \text{cos}(\pi x))$ & 96.0 & 95.8 & 90.5 & 93.6 & \textbf{83.8} & 92.0 \\
         
\rowcolor{blue!10}
         $e^{-x}$  & \textbf{97.0} & \textbf{97.0} & \textbf{90.8} & \textbf{95.2} & 83.5 & \textbf{92.7} \\
    \bottomrule
    \bottomrule
    \end{tabular}}
    \caption{\textbf{Decaying function.} We experiment with different decaying functions in $\mathcal{L}_{pp}$ and observe that exponential decay works best, as denoted in \textbf{bold}. }
    \label{tab:decay}
\end{table}

\paragraph{Effect of curvature.} We empirically evaluate the effect of curvature $c$ in HybridTAS. As summarized in Table \ref{tab:curv}, the model performance is maximized at $c=1$. Except for $c=0.5$, we observe a drop in ``Avg" scores on either side of $c=1$. Low curvatures fail to capture the hierarchy in data, whereas high curvatures can cause numerical instability and distort distances, impacting optimization.

\begin{table}[!h]

\renewcommand{\arraystretch}{1.3}
    \centering
    \resizebox{0.45\textwidth}{!}{
    \begin{tabular}{c|ccc|cc|c}
    \toprule
    \toprule
         \textbf{Curvature} & \textbf{F1@10} & \textbf{F1@25} & \textbf{F1@50} & \textbf{Edit} & \textbf{Acc} & \textbf{Avg}  \\
    \midrule
\rowcolor{blue!10}
         0.1 & 95.8 & 95.0 & 89.7 & 92.7 & 82.8 & 91.2  \\
         0.3 & 93.6 & 93.6 & 89.1 & 92.8 & 81.7 & 90.2 \\
         
\rowcolor{blue!10}
         0.5 & 95.8 & 95.8 & \textbf{91.9} & \textbf{95.6} & 82.8 & 92.4 \\
         0.7 & 93.1 & 93.1 & 87.8 & 90.0 & 81.5 & 89.1 \\
\rowcolor{blue!10}
         0.9 & 95.1 & 95.1 & 90.5 & 92.1 & \textbf{83.7} & 91.3 \\
         1.0 & \textbf{97.0} & \textbf{97.0} & 90.8 & 95.2 & 83.5 & \textbf{92.7} \\
\rowcolor{blue!10}
         2.0 & 95.1 & 95.1 & 91.2 & 93.1 & 82.0 & 91.3 \\
    \bottomrule
    \bottomrule
    \end{tabular}}
    \caption{\textbf{Effect of curvature.} HybridTAS exhibits maximal performance for $c=1$ on the GTEA dataset \cite{gtea}. Best results have been \textbf{bolded}.}
    \label{tab:curv}
\end{table}

\paragraph{Need for two-phase optimization.} During the \textit{Guidance Phase}, the diffusion model corrects its trajectory by aligning the denoised embeddings with the geodesic between the origin and the action prototypes. This requires the action prototypes to be static, as in a two-step optimization strategy. In single-step optimization, the action prototypes behave as dynamic targets, thereby making training unstable. Our experiments further validate this, as shown in Table \ref{tab:optim}.

\begin{table}[!h]

\renewcommand{\arraystretch}{1.3}
    \centering
    \resizebox{0.45\textwidth}{!}{
    \begin{tabular}{c|ccc|cc|c}
    \toprule
    \toprule
         \textbf{Method} & \textbf{F1@10} & \textbf{F1@25} & \textbf{F1@50} & \textbf{Edit} & \textbf{Acc} & \textbf{Avg}  \\
    \midrule
\rowcolor{blue!10}
         One-step optimization & 93.2 & 93.0 & 90.7 & 89.1 & 83.3 & 89.9\\
         Two-step optimization & \textbf{97.0} & \textbf{97.0} & \textbf{90.8} & \textbf{95.2} & \textbf{83.5} & \textbf{92.7} \\
    \bottomrule
    \bottomrule
    \end{tabular}}
    \caption{\textbf{Optimization strategies}. The two-step optimization strategy enables HybridTAS to iteratively refine its trajectory toward static targets (i.e., action prototypes), leading to improved performance over the one-step approach on the GTEA dataset \cite{gtea}. Best results have been \textbf{bolded}.}
    \label{tab:optim}
\end{table}

\paragraph{Effect of training losses.} We conduct ablations exclusively on the hyperbolic loss components, as the cross-entropy loss ($\mathcal{L}_\text{ce}$) is fundamental to our framework. Our findings indicate that HybridTAS attains peak performance when all the proposed hyperbolic losses are jointly employed, highlighting their complementary contributions as shown in Table \ref{tab:ablation_losses}.

\begin{table}[!h]
\renewcommand{\arraystretch}{1.3}
    \centering
    \resizebox{0.45\textwidth}{!}{
    \begin{tabular}{cccc|ccc|cc|c}
    \toprule
    \toprule
        $\mathcal{L}_\text{entail}$ & $\mathcal{L}_\text{margin}$ & $\mathcal{L}_\text{pp}$ & $\mathcal{L}_\text{gg}$ &\textbf{F1@10} & \textbf{F1@25} & \textbf{F1@50} & \textbf{Edit} & \textbf{Acc} & \textbf{Avg}  \\
         
    \midrule
\rowcolor{blue!10}
          \checkmark & &\checkmark & \checkmark  & 95.8 & 95.8 & 91.2 & 94.6 & 83.6 & 92.2 \\
           & \checkmark & \checkmark & \checkmark & 95.0 & 94.2 & 88.8 & 93.2 & 82.3  & 90.7 \\

\rowcolor{blue!10}
\checkmark & \checkmark & \checkmark &  & 95.4 & 95.4 & 88.6 & 92.6 & 82.6 & 90.9 \\
          \checkmark & \checkmark & \checkmark & \checkmark &  \textbf{97.0} & \textbf{97.0} & \textbf{90.8} & \textbf{95.2} & \textbf{83.5} & \textbf{92.7}\\

    \bottomrule
    \bottomrule
    \end{tabular}}
    \caption{\textbf{Effect of training losses.} HybridTAS performs best with all hyperbolic losses, on the GTEA dataset \cite{gtea}. Best results have been \textbf{bolded}.}
    \label{tab:ablation_losses}
\end{table}
\paragraph{Effect of inference steps.} Based on our experiments with different numbers of inference steps, as reported in Table \ref{tab:inf_steps}, we observe a steady and marginal improvement in performance as step number increases. Interestingly, we also observe that HybridTAS outperforms ActFusion \cite{ActFusion} (25 inference steps) with 66\% fewer step numbers. This can be attributed to the improved semantic and hierarchical understanding of the model in the latent space.
\begin{table}[!h]
\renewcommand{\arraystretch}{1.3}
    \centering
    \resizebox{0.45\textwidth}{!}{
    \begin{tabular}{c|ccc|cc|c}
    \toprule
    \toprule
         \textbf{Inference Steps} & \textbf{F1@10} & \textbf{F1@25} & \textbf{F1@50} & \textbf{Edit} & \textbf{Acc} & \textbf{Avg}  \\
    \midrule
\rowcolor{blue!10}
         1 & 83.6 & 83.6 & 78.4 & 75.3 & 83.0 &  80.8\\
         2 & 87.6 & 87.6 & 82.1 & 81.1 & 84.1 & 84.5\\
\rowcolor{blue!10}
         4 & 93.0 & 93.0 & 87.1 & 88.7 & \textbf{84.2} & 89.2\\
         8 & 95.5 & 95.5 & 89.4 & 93.8 & 83.0 & 91.4\\
\rowcolor{blue!10}
         12 & 95.6 & 95.6 & 90.3 & 93.9 & 83.4 & 91.8\\
         16 & 96.1 & 96.1 & 90.1 & 94.4 & 83.8 & 92.1\\
\rowcolor{blue!10}
         20 & 94.5 & 94.5 & 90.4 & 95.3 & 84.0 & 92.6\\
         25 &  \textbf{97.0} & \textbf{97.0} & 90.8 & 95.2 & 83.5 & 92.7\\
\rowcolor{blue!10}
         50 & 96.2 & 96.0 & 92.2 & \textbf{96.3} & 83.7 & 92.9\\
         100 & 96.2 & 96.1 & \textbf{92.4} & \textbf{96.3} & 83.9 & \textbf{93.0}\\
    \bottomrule
    \bottomrule
    \end{tabular}}
    \caption{\textbf{Inference steps.} We ablate on the number of inference steps and observe that HybridTAS surpasses ActFusion \cite{ActFusion} with 66\% fewer steps (8 inference steps) on the GTEA dataset \cite{gtea}. Best results have been \textbf{bolded}.}
    \label{tab:inf_steps}
\end{table}
\section{Quantitative Analysis}

\noindent Table~\ref{tab:segmentation_results} presents the performance of \textbf{HybridTAS} compared to SOTA methods across three benchmark datasets. Our method consistently outperforms prior approaches across all metrics, with particularly substantial improvements on GTEA~\cite{gtea}. On 50Salads~\cite{50salads}, HybridTAS achieves a remarkable improvement of \textbf{+3.6} in F1@50 and \textbf{+3.4} in Edit score over ActFusion~\cite{ActFusion}, highlighting its effectiveness in enhancing temporal consistency and boundary precision. Our average score of \textbf{90.6}, a \textbf{+2.1} gain over baselines, validates the complementary benefit of jointly optimizing Euclidean and hyperbolic objectives. On Breakfast~\cite{breakfast}, we observe notable gains in F1@50 (\textbf{+3.4}) and Accuracy (\textbf{+3.8}), indicating that HybridTAS significantly reduces prediction uncertainty. Improvements in Edit score (\textbf{+1.8}) and the overall average score (\textbf{+2.5}) further demonstrate our model’s ability to capture fine-grained actions. The most pronounced improvements are observed on GTEA~\cite{gtea}, with HybridTAS surpassing ActFusion by \textbf{+3.9}, \textbf{+3.7}, and \textbf{+3.9} in F1@10, F1@25, and F1@50 respectively, and by \textbf{+3.6} in Edit. These gains, leading to an average improvement of \textbf{+3.1}, underscore the efficacy of our hierarchical supervision in modeling atomic human actions with high fidelity.

\begin{figure}[!h]
    \centering
    \includegraphics[width=0.46\textwidth]{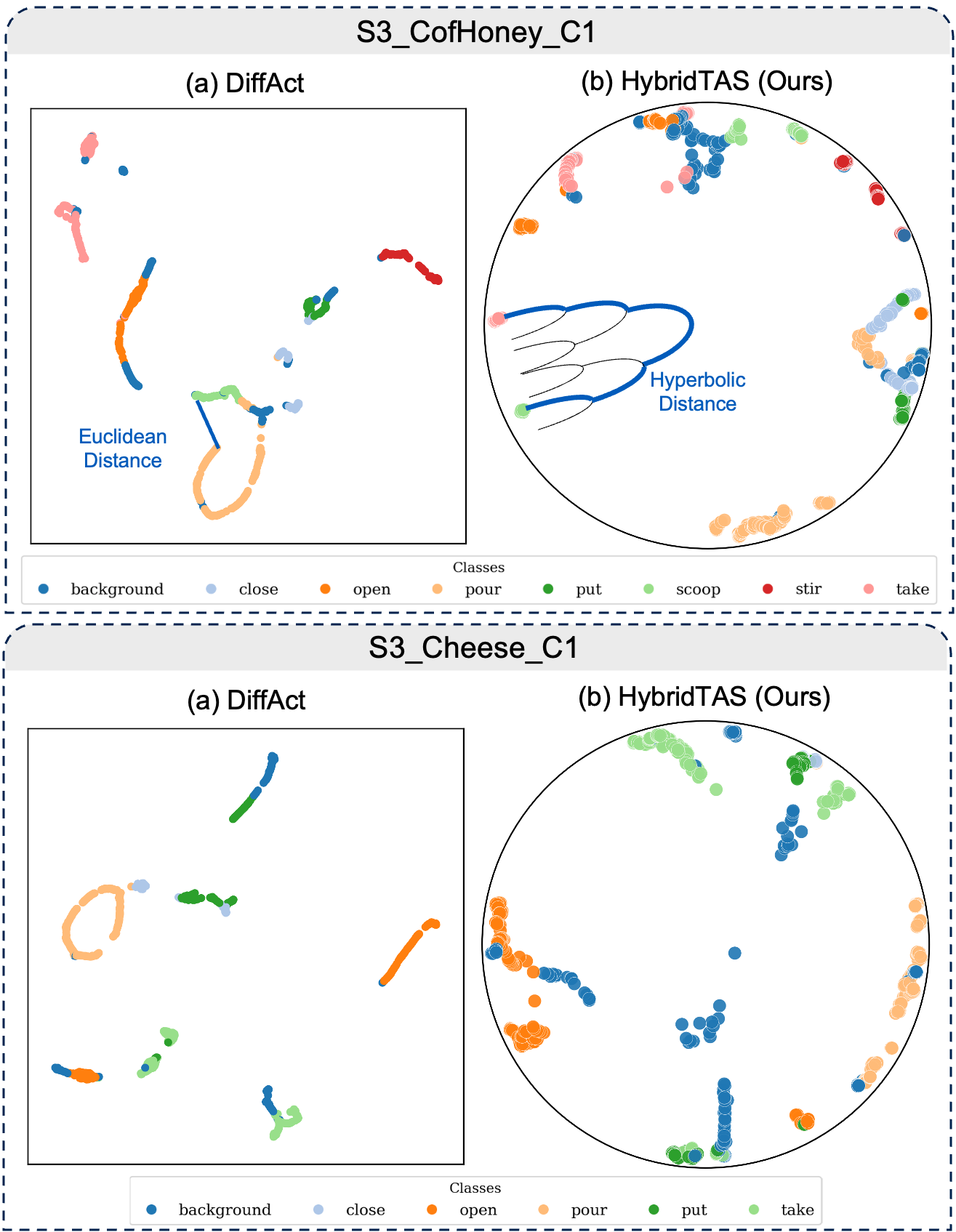}
    \caption{\textbf{Euclidean and Hyperbolic UMAP on the GTEA dataset \cite{gtea}.} UMAP projections of action embeddings from DiffAct \cite{DiffAct} in Euclidean space (left) and HybridTAS (Ours) in hyperbolic space (right). Hyperbolic embeddings yield well-separated clusters, with background states centralized and specific actions arranged radially. Further, it better preserves inter-class boundaries and highlights hierarchical structure across tasks. Refer to Fig. \ref{fig:cluster_dist} (Appendix) for cluster centroid specific distances.}
    \label{fig:gtea_umap1}
\end{figure} 

\begin{figure}[!h]
    \centering
    \includegraphics[width=0.47\textwidth]{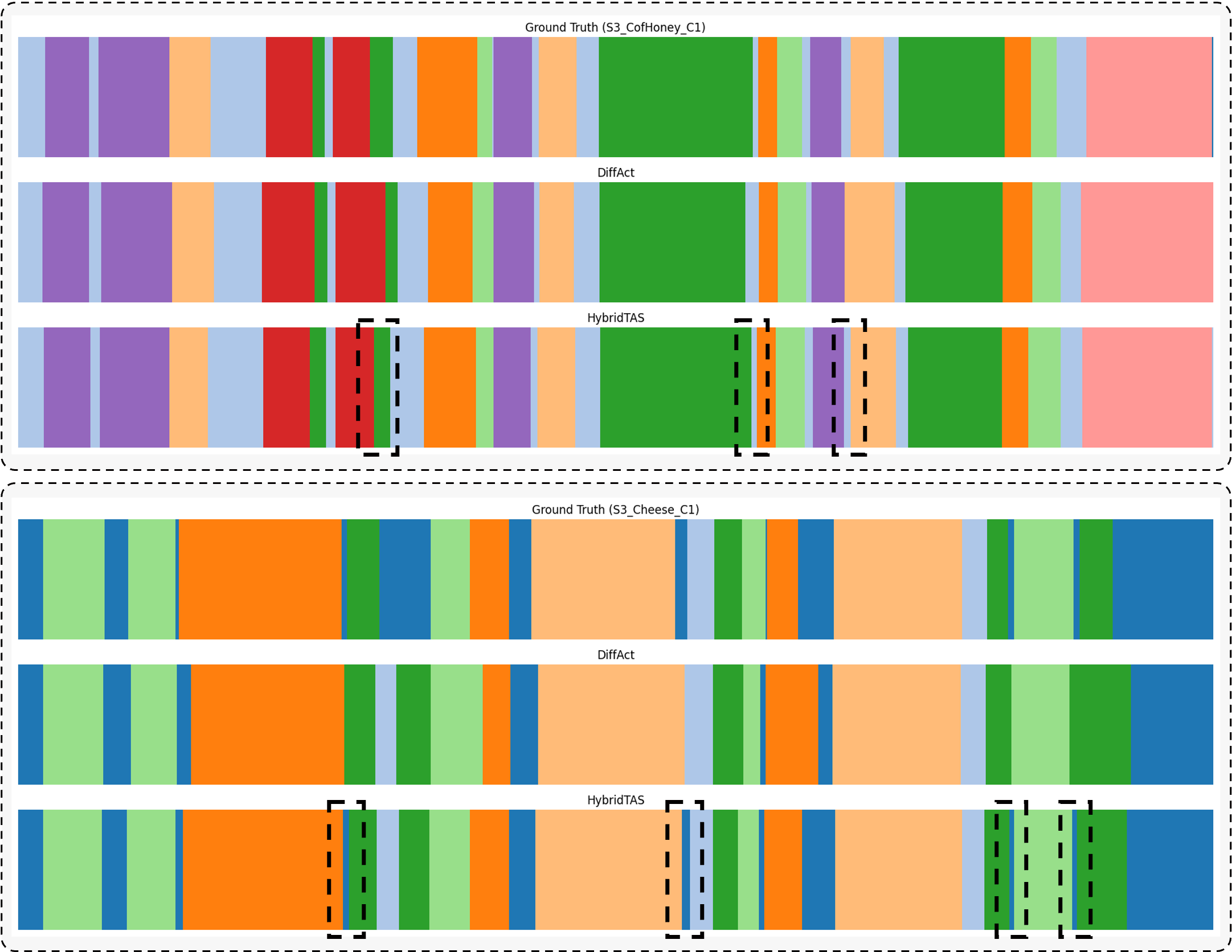}
    \caption{\textbf{Qualitative results on the GTEA dataset \cite{gtea}.} We present a comparison of segmentation outputs of DiffAct \cite{DiffAct} and HybridTAS (Ours) on \textit{S3\_CofHoney\_C1} (top) and \textit{S3\_Cheese\_C1} (bottom) with dashed boxes representing areas of improvement. HybridTAS yields clearer temporal boundaries, better preserves short actions, and maintains semantic consistency across transitions.}
    \label{fig:gtea_samples}
\end{figure} 
\paragraph{Computational cost.} Hyperbolic projections introduce a marginal increase in training time compared to DiffAct \cite{DiffAct}. On the other hand, inference time remains unchanged as we have leveraged 25 inference steps, similar to previous works. 

\section{Qualitative Analysis}
\noindent Fig.~\ref{fig:gtea_umap1} presents 2D UMAP projections of action embeddings learned by DiffAct \cite{DiffAct} and HybridTAS (Ours). For each of the videos, we compare the Euclidean projection (left) with the hyperbolic projection (right). In both tasks (\textit{Cheese} and \textit{CofHoney}), the hyperbolic embedding produces clearer inter-class boundaries compared to the Euclidean counterpart. For instance, in \textit{S3-Cheese-C1}, the actions \textit{open}, \textit{pour}, and \textit{take} appear as compact, radially separated clusters in the hyperbolic space, whereas their Euclidean projections exhibit elongated and partially overlapping trajectories. \textbf{This observation suggests that hyperbolic geometry better preserves the hierarchical separability of action classes and improves boundary definition.} \textbf{The Poincaré embeddings emphasize cluster compactness, with same-class samples pulled toward localized regions near the boundary.} This effect is particularly visible for the \textit{take} and \textit{open} classes, which are tightly grouped in hyperbolic space but more diffusely scattered in Euclidean space. Moreover, hyperbolic spaces implicitly encode a hierarchy: high-frequency background states are centralized, while task-specific actions (e.g., \textit{open}, \textit{pour}, \textit{scoop}) occupy peripheral zones, reflecting their relative semantic specificity. When comparing Cheese and CofHoney tasks, we observe consistent structuring of common classes (\textit{open}, \textit{take}, \textit{pour}). \textbf{Despite contextual differences in the tasks, the hyperbolic model projects these actions into analogous radial sectors, suggesting that the representation generalizes across recipes while retaining class separability.} In contrast, Euclidean projections show more task-specific drifts, particularly for the \textit{background} and \textit{put} actions. More samples have been provided in the appendix.

Figure~\ref{fig:gtea_samples} compares the segmentation outputs of DiffAct \cite{DiffAct} with HybridTAS (Ours), alongside ground truth annotations. In both sequences, HybridTAS demonstrates tighter alignment with ground truth action boundaries compared to DiffAct. Dashed regions highlight instances where DiffAct fails to capture short actions or incorrectly models action boundaries, whereas HybridTAS more faithfully captures transitions between segments. For example, in Cheese, HybridTAS captures short \textit{background} (blue) transitions unlike DiffAct.

\section{Conclusion}
In this paper, we introduce \textbf{HybridTAS}, a novel diffusion-based framework for temporal action segmentation that integrates both Euclidean and hyperbolic geometries to capture the hierarchical structure of actions. By modeling the denoising process along semantically meaningful trajectories in hyperbolic space, HybridTAS enables coarse-to-fine action latent generation. Our two-phase training strategy ensures temporally coherent predictions and provides the model with targets to correct its denoising trajectory. Extensive experiments on GTEA, 50Salads, and Breakfast datasets confirm that HybridTAS outperforms existing diffusion-based models across all standard metrics. Beyond improved segmentation, our approach enables faster convergence in fewer inference steps due to its geometry-aware denoising path. 

{\small
\bibliographystyle{ieee_fullname}
\bibliography{egbib}
}

\newpage
\appendix
\section{Technical Appendices and Supplementary Material}
\subsection{DiffAct: Diffusion Action Segmentation}
\label{subsec:DiffAct}
\noindent DiffAct \cite{DiffAct} introduces a novel generative approach for temporal action segmentation by leveraging denoising diffusion probabilistic models (DDPMs). Unlike prior methods that operate deterministically, DiffAct formulates action segmentation as a conditional generation problem, where frame-wise action sequences are generated from pure noise conditioned on video features. In this paper, we adopt DiffAct's model architecture and input masking strategies during training. 

\paragraph{Diffusion-based formulation.} Given an input video with $L$ frames and corresponding ground truth one-hot action labels $Y_0 \in \{0,1\}^{L \times C}$ (where $C$ is the number of action classes), and an encoder $h_\phi$.
The encoder encodes the input video features $F \in \mathbb{R}^{L \times D}$ using $E = h_\phi(F)$. A decoder $g_\psi$ is trained to denoise the noisy label sequence $Y_t$ at timestep $t$ conditioned on encoded features $E$, producing action logits $P_t \in \{0,1\}^{L \times C}$.

\paragraph{Training.} Beyond proposing novel euclidean training objectives, DiffAct uses a condition masking strategy rooted in human behavior modelling. Specifically, they integrate three human action priors into the diffusion framework. Firstly, \textit{No Masking}, which passes all features into the decoders. Secondly, \textit{Masking for Position Prior} and \textit{Masking for Boundary Prior} to enforce the model to rely only on frame positions and explore action boundaries. Lastly, \textit{Masking for Relation Prior} prompts the model to infer the missing action segment.   
\paragraph{Inference.} The denoising decoder $g_\psi$ is trained to handle inputs with varying levels of noise, even sequences composed entirely of random noise. During inference, the process begins with a purely noisy sequence $\hat{Y}_T \sim \mathcal{N}(0, I)$ and gradually removes the noise through an iterative denoising procedure. At each step $t$, the sequence is updated using:

{\small
\begin{equation}
\hat{Y}_{t-1} = \sqrt{\bar{\alpha}_{t-1}} P_t 
+ \frac{\sqrt{1 - \bar{\alpha}_{t-1} - \sigma_s^2}}{\sqrt{1 - \bar{\alpha}_t}} (\hat{Y}_t 
- \sqrt{\bar{\alpha}_s} P_t) 
+ \sigma_t \epsilon
\end{equation}
}
where $\hat{Y}_{t-1}$ is passed into the decoder to produce the next prediction $P_{t-1}$. This process continues step-by-step, refining the noisy sequence 
$\hat{Y}_T, \hat{Y}_{T-1}, \ldots, \hat{Y}_0$ 
until the final output $\hat{Y}_0$, which closely approximates the true action sequence.

To accelerate inference, DiffAct adopts a sampling trajectory that skips intermediate steps, producing a shorter sequence such as 
$\hat{Y}_S, \hat{Y}_{S-\Delta}, \ldots, \hat{Y}_0$. 
Note that during inference, the encoded features $E$ are fed into the decoder without any masking.

\subsection{Background on Diffusion Models}
\label{sec:appendix_bg}
 \noindent Diffusion models learn to approximate a target data distribution by progressively corrupting data with Gaussian noise in a forward process, and then learning to reverse this corruption through a denoising neural network. The forward (or diffusion) process transforms clean data \( \mathbf{x}_0 \) into a noisy version \( \mathbf{x}_t \) by gradually adding Gaussian noise according to a predefined variance schedule. Specifically, this process can be expressed as:
\begin{equation}
    \mathbf{x}_t = \sqrt{\gamma(t)} \, \mathbf{x}_0 + \sqrt{1 - \gamma(t)} \, \boldsymbol{\epsilon}, \quad \boldsymbol{\epsilon} \sim \mathcal{N}(0, \mathbf{I})
\end{equation}
where \( \gamma(t) \) is a monotonically decreasing function that controls the noise magnitude at timestep \( t \in \{1, 2, \dots, T\} \).

In the reverse process, a neural network \( f(\mathbf{x}_t, t) \) is trained to recover \( \mathbf{x}_0 \) from noisy inputs \( \mathbf{x}_t \). This is typically done by minimizing a simple L2 reconstruction loss:
\begin{equation}
    \mathcal{L} = \frac{1}{2} \left\| f(\mathbf{x}_t, t) - \mathbf{x}_0 \right\|_2^2
\end{equation}
At inference time, the model starts from a pure noise vector \( \mathbf{x}_T \) and iteratively denoises it through the learned reverse trajectory \( \mathbf{x}_T \rightarrow \mathbf{x}_{T-\Delta} \rightarrow \dots \rightarrow \mathbf{x}_0 \), ultimately reconstructing a sample from the original data distribution.

In our setting, the model learns to generate frame-wise action label sequences from Gaussian noise, conditioned on video features for action segmentation.

\subsection{Additional Dataset}
To validate our framework beyond cooking datasets, we utilize the YouTube Instructional (YTI) dataset \cite{yti}. The dataset consists of five tasks and thirty videos per task with an average video duration of two minutes. The data is coarsely labeled on 49 action categories. In Table \ref{tab:yti_results}, we evaluate DiffAct \cite{yti} and HybridTAS (Ours) on this dataset using the same evaluation metrics. Our proposed approach outperforms DiffAct across all metrics. 

\begin{table}[!h]

\renewcommand{\arraystretch}{1.3}
    \centering
    \resizebox{0.45\textwidth}{!}{
    \begin{tabular}{c|ccc|cc|c}
    \toprule
    \toprule
         \textbf{Method} & \textbf{F1@10} & \textbf{F1@25} & \textbf{F1@50} & \textbf{Edit} & \textbf{Acc} & \textbf{Avg}  \\
    \midrule
\rowcolor{blue!10}
         DiffAct \cite{DiffAct} & 53.4 & 45.5 & 27.5  & 56.5 & 71.1 & 50.8 \\
         HybridTAS (Ours) & \textbf{58.1} & \textbf{52.3} & \textbf{33.6} & \textbf{62.3} & \textbf{69.5} & \textbf{54.9}\\
    \bottomrule
    \bottomrule
    \end{tabular}}
    \caption{\textbf{Quantitative Results on the YTI dataset}.}
    \label{tab:yti_results}
\end{table}

\begin{figure}[!h]
    \centering
    \includegraphics[width=0.46\textwidth]{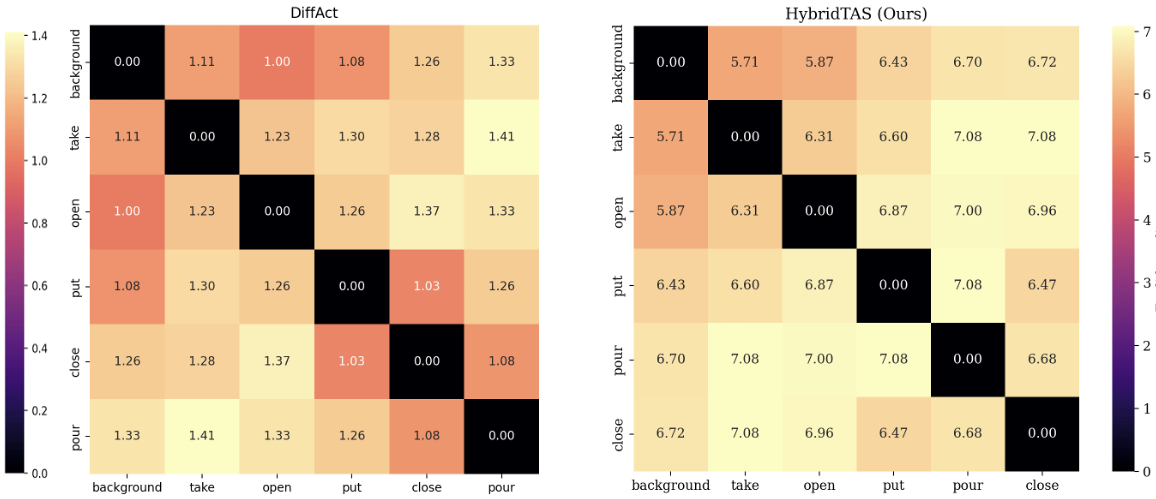}
    \caption{\textbf{Cluster centroid distances.} We plot the cluster centroid distances to showcase an almost 6x more distance in HybridTAS, which is indicative of better clustering.  Note that the HybridTAS distances are hyperbolic distances, whereas DiffAct \cite{DiffAct} distances are Euclidean.}
    \label{fig:cluster_dist}
\end{figure}

\begin{figure}[!h]
    \centering
    \includegraphics[width=0.46\textwidth]{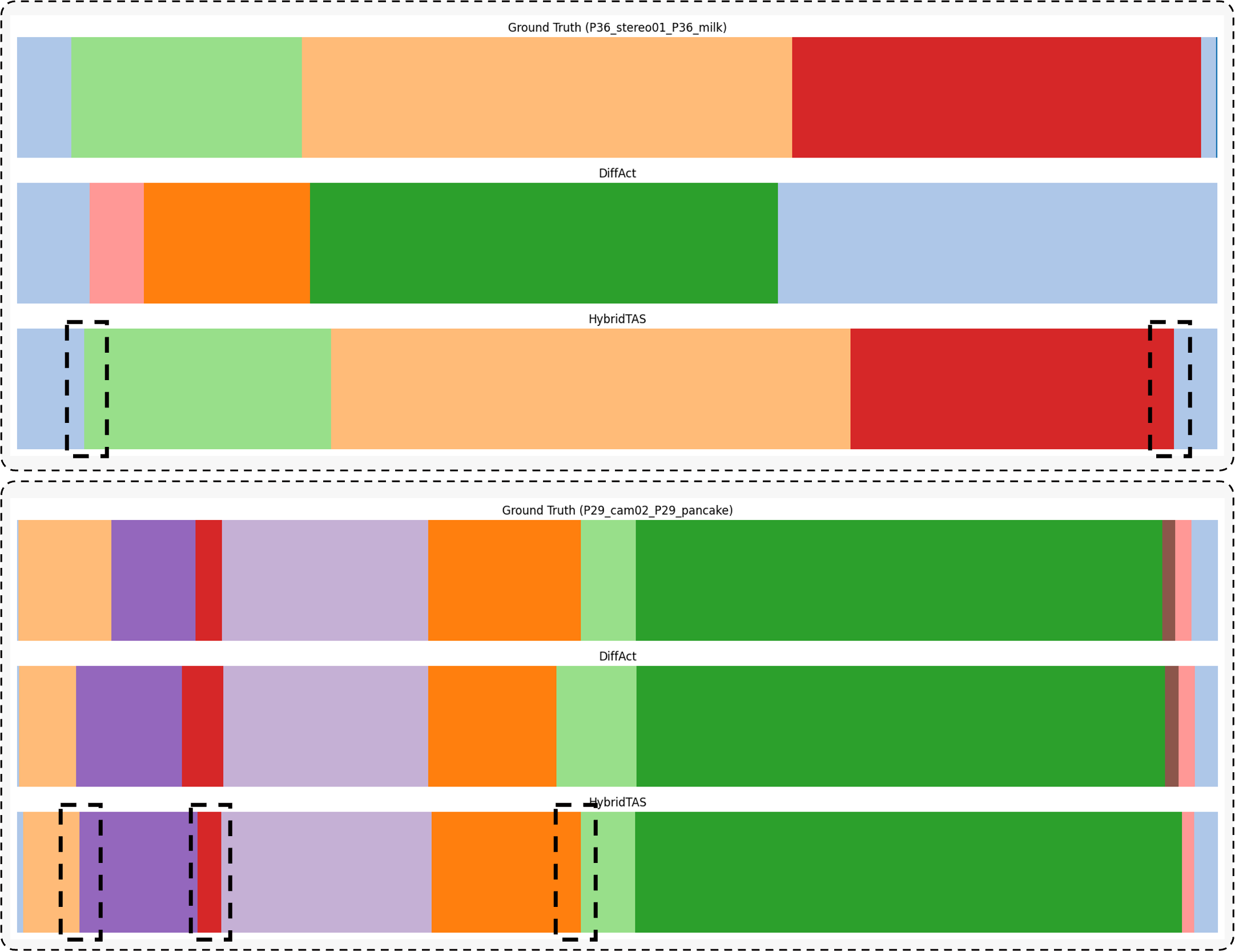}
    \caption{\textbf{Qualitative results on the Breakfast dataset \cite{breakfast}.}We present a comparison of segmentation outputs of DiffAct \cite{DiffAct} and HybridTAS (Ours) on \textit{P30\_stereo01\_P36} (top) and \textit{P29\_cam02\_P29\_pancake} (bottom) with dashed boxes representing areas of improvement.}
    \label{fig:breakfast_samples}
\end{figure} 

\begin{figure}[!h]
    \centering
    \includegraphics[width=0.46\textwidth]{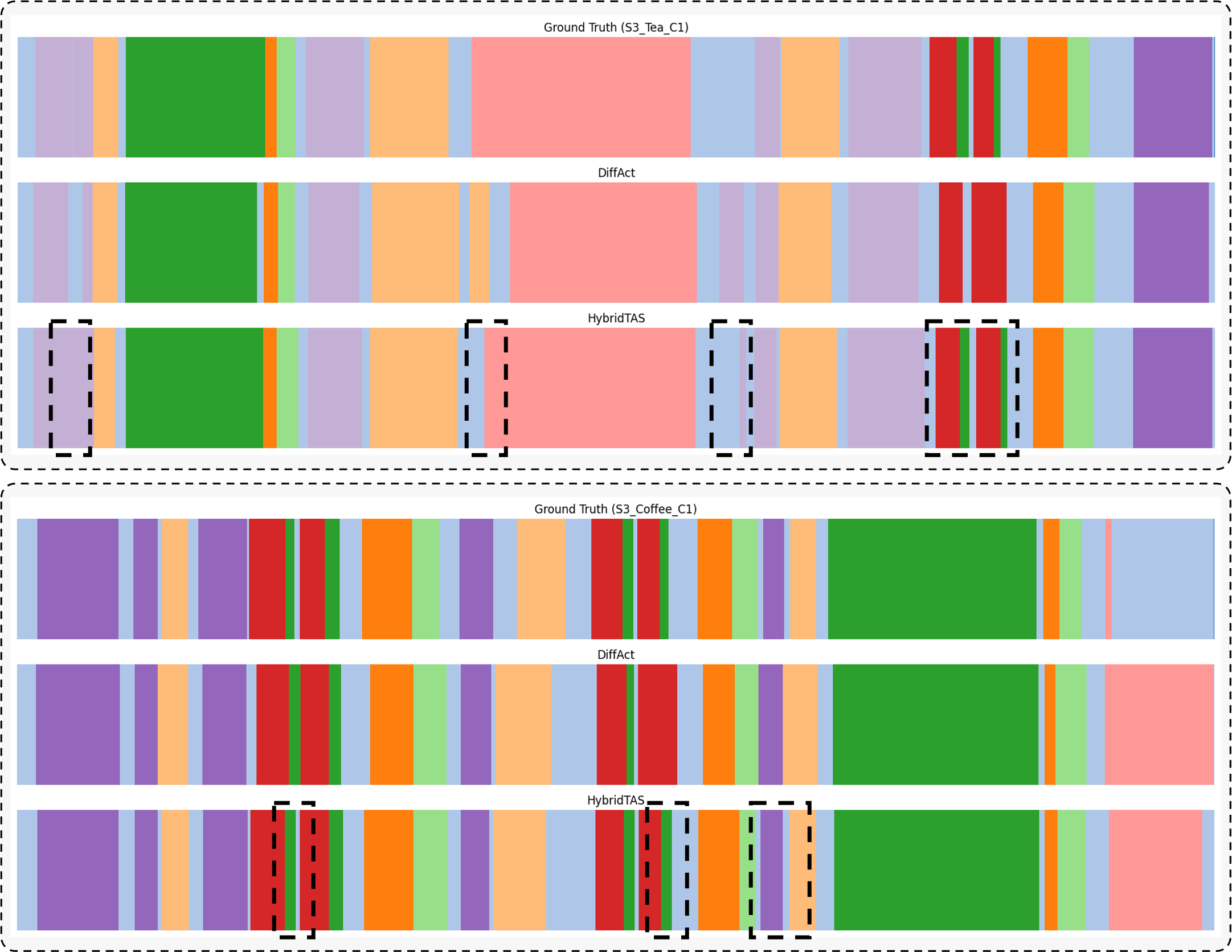}
    \caption{\textbf{Qualitative results on the GTEA dataset \cite{gtea}.}We present a comparison of segmentation outputs of DiffAct \cite{DiffAct} and HybridTAS (Ours) on \textit{S3\_Tea\_C1} (top) and \textit{S3\_Coffee\_C1} (bottom) with dashed boxes representing areas of improvement.}
    \label{fig:gtea_samples2}
\end{figure} 

\begin{figure}[!h]
    \centering
    \includegraphics[width=0.46\textwidth]{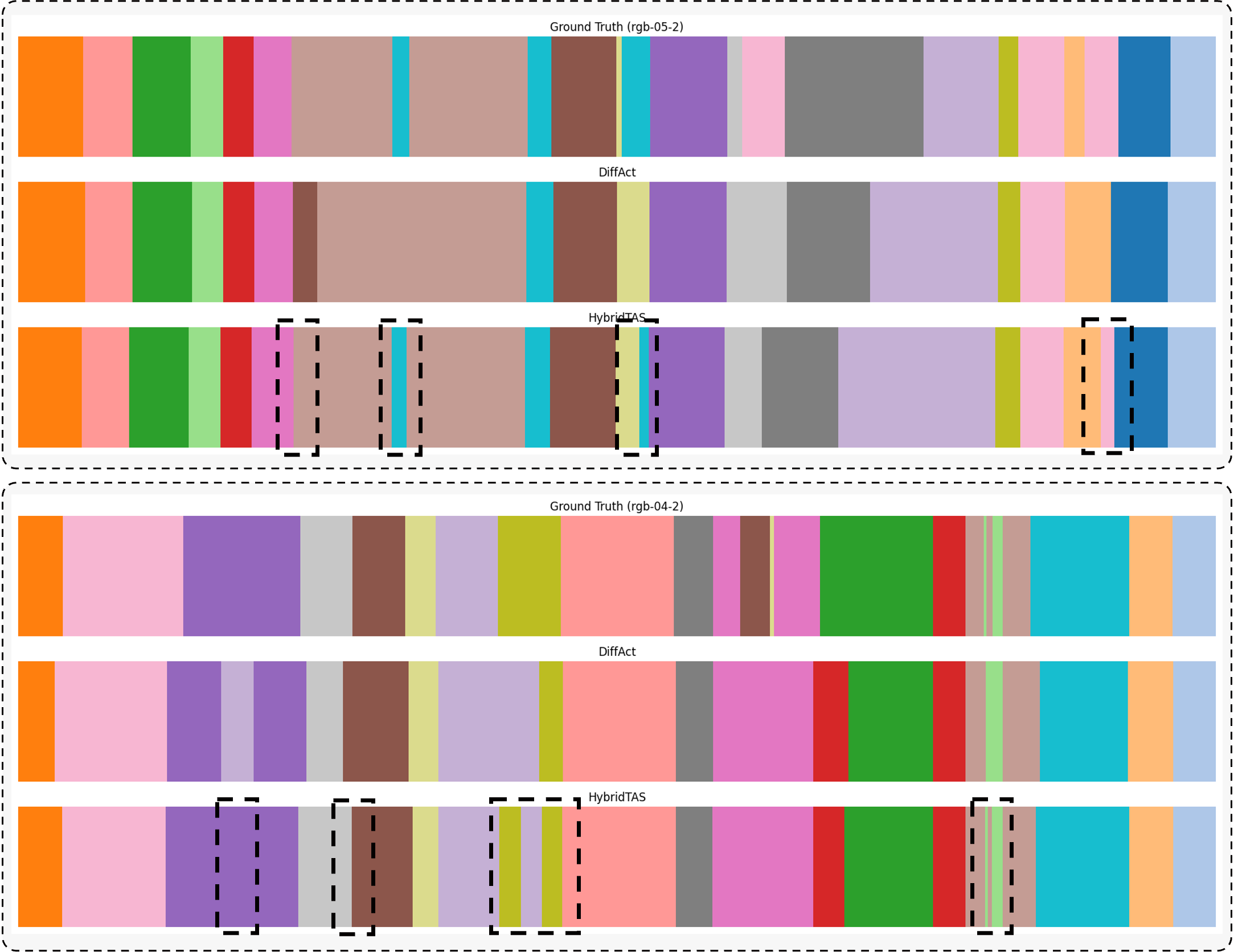}
    \caption{\textbf{Qualitative results on the 50Salads dataset \cite{50salads}.}We present a comparison of segmentation outputs of DiffAct \cite{DiffAct} and HybridTAS (Ours) on \textit{rgb-05-2} (top) and \textit{rgb-04-2} (bottom) with dashed boxes representing areas of improvement.}
    \label{fig:50salads_samples}
\end{figure}

\begin{table}[!h]
\renewcommand{\arraystretch}{1.3}
    \centering
    \resizebox{0.45\textwidth}{!}{
    \begin{tabular}{c|ccc}
    \toprule
    \toprule
         \textbf{Hyperparamters} & \textbf{50 Salads}\cite{50salads} & \textbf{Breakfast} \cite{breakfast} & \textbf{GTEA} \cite{gtea} \\
    \midrule
\rowcolor{blue!10}
         $\lambda_{\text{ce}}$ & 0.5 & 0.5 & 0.5 \\
         $\lambda_{\text{entail}}$ & 0.05 & 0.1 & 0.05 \\
\rowcolor{blue!10}
         $\lambda_{\text{margin}}$ & 0.1 & 0.2 & 0.1 \\
         $\lambda_{\text{pp}}$ & 0.1 & 0.2 & 0.1 \\
\rowcolor{blue!10}
         $\lambda_{\text{gg}}$ & 0.1 & 0.2 & 0.1 \\
         $E_1$ & 2000 & 400 & 4000 \\
\rowcolor{blue!10}
         Curvature ($c$) & 1.0 & 1.0 &  1.0 \\
         Total epochs & 5000 & 1000 & 10000 \\
    \bottomrule
    \bottomrule
    \end{tabular}}
    \caption{\textbf{Dataset specific hyperparamter values.} }
    \label{tab:hyperparams}
\end{table}

\section{Experiments}
\paragraph{Datasets.}We conduct experiments on three benchmark datasets: GTEA, 50Salads, and Breakfast. GTEA \cite{gtea} consists of 28 egocentric videos of daily activities, covering 11 action classes. Each video is approximately one minute long and contains around 19 action instances. 50Salads \cite{50salads} features 50 top-view videos of salad preparation, annotated with 17 action classes. The videos average six minutes in length, with roughly 20 action instances per video. Breakfast \cite{breakfast} is a large-scale dataset comprising 1712 third-person videos spanning 48 action classes related to breakfast preparation. While the average video length is two minutes, there is significant variance across samples; each video contains around seven action instances on average. Among the three, Breakfast \cite{breakfast} offers the largest scale, while 50Salads \cite{50salads} includes the longest videos and the highest number of instances per video. As in DiffAct \cite{DiffAct}, we adopt five-fold cross-validation on 50Salads and four-fold cross-validation on GTEA and Breakfast, using the same data splits for fair comparison.

\paragraph{Metrics.}Following previous works \cite{MSTCN++, ASFormer}, the frame-wise accuracy (Acc), the edit score (Edit), and the F1 scores at overlap thresholds 10\%, 25\%, 50\% (F1@{10, 25, 50}) are reported. The accuracy assesses the results at the frame level, while the edit score and F1 scores measure the performance at the segment level.

\paragraph{Implementation details.} For all datasets, we utilize the I3D features \cite{I3D} as the input features $\mathbf{F}$, whose dimension is 2048. The encoder $h_\phi$ and decoder $g_\psi$ are adopted from DiffAct \cite{DiffAct}. The encoder is a reimplementation of the ASFormer encoder \cite{ASFormer}, while the ASFormer decoder is modified to be step-aware by incorporating step embeddings into the input, as proposed in \cite{ho2020denoisingdiffusionprobabilisticmodels}. Specifically, the encoder contains 10, 10, 12 layers with 64, 64, 256 feature maps for the GTEA \cite{gtea}, 50Salads \cite{50salads}, and Breakfast \cite{breakfast} datasets. The decoder comprises of 8 layers with 24, 24, 128 feature maps for the respective datasets. Intermediate features from three encoder layers (5, 7, 9) are concatenated to be used as conditional input to the decoder. The entire framework is trained with the RiemannianAdam optimizer, a batch size of 4, a learning rate of $1e-4$ (Breakfast \cite{breakfast}) and $5e-4$ (GTEA \cite{gtea} and 50Salads \cite{50salads}. The total diffusion timesteps during training is set to $T=1000$, and 25 steps are utilized during inference. We have performed all experiments on a single NVIDIA H100 GPU. Dataset-specific hyperparameters have been provided in Table \ref{tab:hyperparams}.

\end{document}